\documentclass[journal]{IEEEtran}
\usepackage{amsmath,amsfonts}
\usepackage{amssymb}
\usepackage{algorithm}
\usepackage[indLines=true,noEnd=false]{algpseudocodex}
\usepackage{array}
\usepackage[caption=false,font=normalsize,labelfont=sf,textfont=sf]{subfig}
\usepackage{textcomp}
\usepackage{threeparttable}
\usepackage{booktabs}
\usepackage{graphicx}
\usepackage{float}
\usepackage{stfloats}
\usepackage{url}
\usepackage{verbatim}
\usepackage{cite}
\usepackage{bm}
\usepackage
[colorlinks=true, 
linkcolor=blue, 
citecolor=blue, 
urlcolor=cyan, 
bookmarksopen=true]{hyperref}
\usepackage{xcolor}

\begin{document}

\title{Restoration Flow Matching-Based Channel Refinement and Equalization Correction for MIMO Semantic Communications}

\author{
Wenkai Liu, 
Nan Ma,~\IEEEmembership{Member,~IEEE,} 
Jianqiao Chen,
Xiaodong Xu,~\IEEEmembership{Senior Member,~IEEE,}
Meixia Tao,~\IEEEmembership{Fellow,~IEEE,} 
and 
Ping Zhang,~\IEEEmembership{Fellow,~IEEE} 
        % <-this % stops a space
\thanks{
Wenkai Liu, Nan Ma, Xiaodong Xu and Ping Zhang are with the State Key Laboratory of Networking and Switching Technology, Beijing University of Posts and Telecommunications, Beijing 100876, China 
(e-mail: liuwenkai@bupt.edu.cn; manan@bupt.edu.cn; xuxiaodong@bupt.edu.cn; pzhang@bupt.edu.cn).(Corresponding author: Nan Ma.)

Jianqiao Chen is with the ZGC Institute of Ubiquitous-X Innovation and Applications, Beijing 100876, China (e-mail: jqchen1988@163.com).

Meixia Tao is with the Cooperative Medianet Innovation Center (CMIC) and the Department of Electronic Engineering, Shanghai Jiao Tong University, Shanghai 200240, China (mxtao@sjtu.edu.cn).
}}% <-this % stops a space
% The paper headers
% \markboth{Journal of \LaTeX\ Class Files,~Vol.~14, No.~8, August~2021}%
% {Shell \MakeLowercase{\textit{et al.}}: A Sample Article Using IEEEtran.cls for IEEE Journals}

% Remember, if you use this you must call \IEEEpubidadjcol in the second
% column for its text to clear the IEEEpubid mark.
\maketitle
\begin{abstract}
In multiple-input multiple-output (MIMO) semantic communication, imperfect channel state information (CSI) and equalization mismatch can seriously degrade semantic reconstruction quality.
To address this issue, we propose a unified restoration flow matching (RFM)-based framework for channel refinement and equalization correction. 
Specifically, the channel RFM (CRFM) module is developed to refine the coarse channel, thereby improving channel estimation accuracy.
Based on the refined channel, the developed semantic RFM (SRFM) module is employed to correct the residual distortions in the post-equalization latent space.
The key idea is to formulate the two cascaded inverse problems of channel estimation and equalization as the unified conditional restoration task, in which the learned conditional velocity field guides the perturbed distribution towards the target distribution.
To enhance the robustness of these two modules under various distortion conditions, we develop a dual-anchor perturbation training strategy that jointly learns near-manifold refinement and large-error correction, and implement inference through a few-step deterministic ordinary differential equation (ODE) solver.
Extensive experiments on MIMO channels and visual semantic transmission tasks demonstrate that the proposed scheme improves key metrics for channel estimation and semantic reconstruction quality. 
Moreover, compared with representative diffusion-based generative baselines, the proposed method requires fewer sampling steps.

\end{abstract}

\begin{IEEEkeywords}
Semantic communication, flow matching, channel estimation and equalization, MIMO communications.
\end{IEEEkeywords}

\section{Introduction}
Semantic communication has emerged as a highly promising paradigm for next-generation wireless networks by shifting the design objective from bit-level fidelity to the transmission of task-relevant information and semantic preservation \cite{zhangToward2022}, \cite{YangSemantic2023}, \cite{ChaccourLess2025}. 
This paradigm is particularly appealing for 6G visual transmission services, as the massive scale of source data and complex channel environments make traditional approaches that separate source coding from channel coding increasingly inefficient.
In this context, deep joint source-channel coding (DeepJSCC) has become one of the most representative implementation frameworks for semantic communication, as it enables the end-to-end optimization of semantic representation, channel adaptation, and reconstruction quality within a neural network architecture \cite{Kountouris_Semantic_2021}, \cite{Dai_Nonlinear_2022}.

Following this line of thought, a deep learning-based JSCC framework was proposed in \cite{Bourtsoulatze_Deep_2019}, in which image compression and error correction were jointly realized through a convolutional neural network (CNN)-based autoencoder architecture.
In addition, Transformer-based architectures have demonstrated superior capabilities in modeling high-dimensional visual semantics \cite{Vaswani_Attention_2017}. 
In particular, the hierarchical shifted-window design of the Swin Transformer achieves an ideal balance among multi-scale feature extraction, long-range dependency modeling, and computational efficiency, making it highly suitable for semantic image transmission \cite{Liu_Swin_2021}. 
The performance of these models degrades sharply when channel noise is severe or the testing statistics deviate from the training distribution. 
To address this issue, the authors in \cite{Xu_ADJSCC_2022} proposed an adaptive JSCC with an attention module, exhibiting enhanced robustness under additive white Gaussian noise (AWGN) channel mismatch conditions. 
Subsequently, DeepJSCC explicitly established signal-to-noise ratio (SNR) adaptation as a design objective, emphasizing the robustness of a single model when operating under varying SNR conditions \cite{Ding_SNR_Adaptive_2021}. 
In \cite{Yang_SwinJSCC_2025}, the authors used the Swin transformer as the backbone of the JSCC, and the SNR and rate adaptation techniques were able to adapt to different channel conditions and transmission rates.
These end-to-end scheme studies enable semantic transceivers to adapt to channel variations, but they rely on direct decoding from degraded semantic features, which limits the high-fidelity recovery of source information in the presence of severe noise or channel distribution mismatches.

With the emergence of diffusion models, which excel at generating realistic textures in image synthesis \cite{Ho_DDPM_2020}, various diffusion models are now being applied to improve the quality of reconstructed images, even under high compression ratios \cite{Scalable_2023}, \cite{Welker_DriftRec_2024}. 
This trend has further propelled the development of generative denoising and recovery methods tailored for semantic communication systems \cite{Wu_CDDM_2024}, \cite{Xu_SPEDNSC_2025}, \cite{Wu_ICDM_2026}. 
In wireless semantic communication, the channel denoising
diffusion models (CDDM) is deployed as a denoising module following channel equalization to achieve channel adaptivity in \cite{Wu_CDDM_2024}.
The authors in \cite{Xu_SPEDNSC_2025} proposed an unsupervised latent diffusion posterior sampling method that achieves joint semantic equalization and denoising.
Further, the authors in \cite{Wu_ICDM_2026} proposed an interference cancellation diffusion model (ICDM), which formulates the interference cancellation problem as a maximum a posteriori problem for the joint posterior probability of the signal and interference.
Therefore, under the influence of multiplicative channel distortion, explicit equalization becomes a critical component of the semantic receiver. Its role is to suppress amplitude-phase distortion and feature coupling caused by the channel, thereby providing a latent representation that is easier to recover for subsequent semantic reconstruction.

However, the majority of existing works primarily focus on single-input single-output (SISO) channels \cite{Xu_DeepJSCC_2023}. 
In view of the leading role that MIMO technology plays in enhancing channel capacity and transmission reliability, extending semantic communication from single-antenna links to MIMO systems is critically important \cite{Chataut_MassiveMIMO_2020}.
In MIMO configurations, the received semantic information exhibits the nonlinear mixture and multi-stream coupling induced by the channel matrix, which poses significant challenges to robust semantic transmission. 
To address this issue, several representative works have explored robust transmission mechanisms from various perspectives in the context of MIMO semantic communications \cite{Bian_SpaceTime_MIMO_2023}, \cite{Duan_DMMIMO_2024}, \cite{Wu_DeepJSCC_MIMO_2024}, \cite{Park_ImportanceAware_2026}. 
In \cite{Bian_SpaceTime_MIMO_2023}, orthogonal space-time block coding (OSTBC) was utilized to improve the robustness of image transmission against MIMO channel variations, and an equalizer was employed to estimate the codewords. 
In \cite{Duan_DMMIMO_2024}, the authors performed diffusion-based denoising on the equalized sub-channels. 
Furthermore, the authors developed the DeepJSCC framework, which is specifically tailored for MIMO communications. 
This framework employed a vision transformer-based architecture that leverages both contextual semantic features and channel conditions through self-attention mechanisms \cite{Wu_DeepJSCC_MIMO_2024}.
Another study proposed a communication framework for MIMO-OFDM systems based on semantic importance, which utilizes semantic importance to jointly optimize quantization, subcarrier mapping, and power allocation \cite{Park_ImportanceAware_2026}.

Although the aforementioned studies have improved the recovery quality of corrupted semantic representations and advanced research on semantic communication, they largely rely on the assumption of perfect CSI at the receiver. 
This assumption is often difficult to satisfy in practical wireless environments, thereby limiting the real-world applicability of these methods. 
Therefore, it is crucial to design high-precision, low-complexity channel estimation methods under pilot-constrained conditions.
In fact, channel estimation itself is a classic inverse problem with a wealth of existing literature \cite{Ozdemir_Channel_2007}, \cite{Ruan_OMP_NearField_2024}. 
The least squares (LS) was widely used due to its simplicity, but its estimation accuracy degrades significantly under low SNR conditions  \cite{Ozdemir_Channel_2007}. 
Sparse recovery methods, such as orthogonal matching pursuit (OMP) \cite{Ruan_OMP_NearField_2024}, can leverage channel sparsity to achieve superior performance. 
However, their performance depends on the quality of the sparse representation dictionary and the validity of the sparsity assumption.
In recent years, deep learning (DL) methods have provided new insights into channel estimation by exploiting statistical correlations in wireless channels \cite{Liu_Deep_2022}, \cite{Arvinte_MIMO_2023}, \cite{Zhou_Generative_2025}, \cite{Feng_DiSeLCE_2025}, \cite{Weng_SemanticMIMO_STT_2024}. 
The authors in \cite{Liu_Deep_2022} proposed a denoising CNN based on deep residual learning for channel denoising. 
In \cite{Arvinte_MIMO_2023}, the authors further proposed a score-based method with posterior sampling.
In \cite{Zhou_Generative_2025}, a generative prior-based MIMO channel estimation method was proposed, which can handle quantized observations even when using low-resolution analog-to-digital converters (ADCs).
In the field of semantic communications, a few studies have also begun to explicitly consider the channel estimation module. 
For example, \cite{Feng_DiSeLCE_2025} developed an end-to-end digital semantic communication system and introduced a ResNet-based channel estimator for OFDM transmission over frequency-selective fading channels. 
Additionally,\cite{Weng_SemanticMIMO_STT_2024}  proposed a semantic-aware system tailored for speech-to-text transmission and extended it with a neural network channel estimation module to mitigate the system's reliance on precise CSI.

Nevertheless, existing research still lacks systematic studies on semantic recovery under imperfect-CSI conditions. 
In MIMO semantic communication, the receiver essentially needs to solve two cascaded and coupled inverse problems. 
First, the channel is estimated based on pilot observations, and then equalization is performed using imperfect-CSI to recover the semantic latent state. 
Since channel estimation errors directly lead to equalization mismatch \cite{Y_MIMO_detection} and propagate to subsequent semantic reconstruction modules, the overall semantic recovery performance is significantly degraded. 

Therefore, to solve these cascaded inverse problems of channel estimation and equalization, we transform both into generative model-based restoration tasks.
However, for latency-sensitive communication receivers, a key limitation of diffusion-based generative models is that the sampling process is typically iterative and relatively slow \cite{Song_DDIM_2021}, \cite{Ma_DiffusionModel_2024},\cite{Fesl_DiffusionBased_2024}, \cite{Wu_ICDM_2026}, \cite{Chen_Generative_2025}. 
Compared with standard diffusion sampling, accelerated variants such as DDIM reduce the reverse sampling step\cite{Song_DDIM_2021}, \cite{Ma_DiffusionModel_2024}.
Some studies truncated the denoising chain according to channel conditions or receiver-side SNR information \cite{Fesl_DiffusionBased_2024,Wu_ICDM_2026}. 
In addition, the variational inference method based on score matching was introduced to directly approximate the posterior distribution of the MIMO channel, thereby achieving efficient channel recovery by reducing sampling steps \cite{Chen_Generative_2025}. 
Nevertheless, the inference process generally still requires multiple successive denoising evaluations.
Recently, the flow matching (FM) framework \cite{lipman2023flow}, based on continuous normalizing flows, represents the latest advancement in generative modeling. 
Instead of explicitly simulating step-by-step reverse diffusion on stochastic differential equation (SDE) like traditional diffusion models, FM directly learns a continuous, deterministic transport process from a simple distribution to the target distribution by regressing a time-dependent vector field along a predefined probability path.
More importantly, FM supports deterministic inference based on ordinary differential equation (ODE) and enables fast sampling utilizing standard numerical ODE solvers. 
Recent work \cite{Liu_Flow_2025}, \cite{Fu_Land_2026} has introduced FM to wireless communications, achieving faster inference while maintaining recovery accuracy by learning the vector field of the signal distribution.

Motivated by these observations, we develop a unified RFM-based framework for robust semantic recovery in practical MIMO receivers subject to cascaded distortions from pilot-based CSI acquisition and CSI-dependent equalization. 
The framework formulates channel refinement and equalization correction as two domain-specific yet structurally aligned conditional restoration tasks. 
The main contributions of this work are summarized as follows.

\begin{itemize}
    \item  We formulate the problem of practical MIMO semantic image receiver under imperfect CSI conditions as a cascaded inverse recovery problem.
    This modeling approach elucidates how channel estimation errors caused by pilot signals propagate through the CSI-dependent equalization process and ultimately translate into structured distortion in the semantic latent domain.
    Furthermore, we reveal that channel refinement and equalization correction share a common restoration structure, both recovering the clean target from a disturbed state.

    \item To counteract these two distinct types of interference that affect semantic decoding, we unify channel refinement and equalization correction into a cascaded RFM task.
    We construct optimal transport-based conditional restoration paths in the channel and semantic-latent domains, where controlled corrupted source states are transported toward paired clean target states through velocity fields learned under constant-velocity supervision.
    Specifically, we develop a CRFM module to mitigate channel estimation errors, and a SRFM module to suppress residual latent distortion after equalization.
    We further show that the learned velocity field provides a local restoration direction of the Gaussian-smoothed clean-state distribution.

    \item We propose a dual-anchor perturbation training strategy for learning robust restoration flows under various receiver distortion conditions.
     By learning a shared velocity field through a mixture of low-perturbation and high-perturbation paths, this approach provides an implicit multiscale restoration mechanism that enables the restoration model to handle varying degrees of distortion.
    
    \item Based on the learned restoration velocity field, we propose an efficient cascaded ODE inference scheme.
    Compared with representative diffusion-based generative baselines, the proposed method requires fewer sampling steps.
    Experiments on MIMO channels and visual semantic transmission tasks demonstrate that the proposed scheme achieves superior performance in both channel estimation and semantic reconstruction quality.
\end{itemize}

\subsection{Organization}
The remainder of this paper is organized as follows. 
Section \ref{System model and problem formulation} presents the system model and problem formulation. 
Section \ref{III} develops the cascaded restoration flow matching within the SwinT-JSCC framework. 
Section \ref{Numerical Results} provides numerical results and ablation studies. 
Section \ref{Conclusion} concludes the paper.

\textit{Notation:} 
Scalars, vectors, and matrices are denoted by italic letters, boldface lower-case letters, and boldface upper-case letters, respectively. 
The superscripts $(\cdot)^T$, $(\cdot)^*$, $(\cdot)^H$, $(\cdot)^{-1}$, and $(\cdot)^{\dagger}$ denote transpose, conjugate, Hermitian transpose, inverse, and pseudo-inverse, respectively. 
The operators $\otimes$, $\odot$, $\mathrm{vec}(\cdot)$, and $\mathrm{tr}(\cdot)$ denote Kronecker product, Hadamard product, vectorization, and trace. 
The Euclidean and Frobenius norms are denoted by $\|\cdot\|_2$ and $\|\cdot\|_F$, respectively. 
Moreover, \(\lceil \cdot \rceil\) denotes the ceiling function, and $\mathbb E[\cdot]$ denotes expectation. 
Finally, $\mathcal N(\bm{\mu},\bm{\Sigma})$ and $\mathcal{CN}(\bm{\mu},\bm{\Sigma})$ denote real Gaussian and circularly symmetric complex Gaussian distributions, respectively.

\section{System model and problem formulation}
\label{System model and problem formulation}
\subsection{Semantic Transmitter Design}
\begin{figure*}[!t]
\centering
\includegraphics[width=6in]{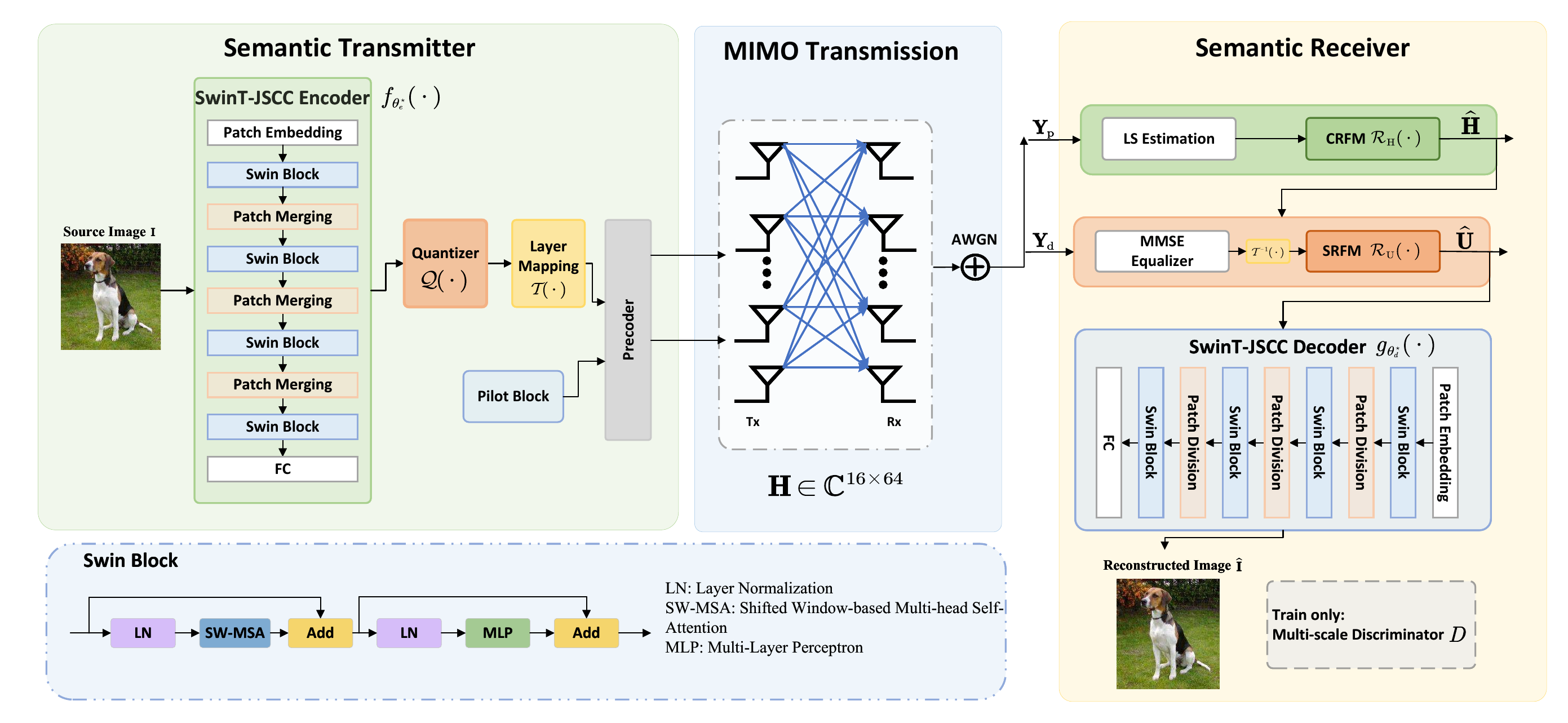}
\caption{Overall architecture of the cascaded RFM-based MIMO semantic communication system.}
\label{Architecture}
\end{figure*}

As illustrated in Fig.~\ref{Architecture}, we consider a visual semantic communication system over a MIMO channel \(\mathbf H\in\mathbb C^{N_r\times N_t}\) with \(N_t\) transmit antennas and \(N_r\) receive antennas.
The transmitter sends \(N_s\) spatial streams, where \(N_s\leq \min\{N_t,N_r\}\).
The transmitted frame is composed of a pilot block followed by a semantic data block.
The pilot block is represented by
$
\mathbf P\in\mathbb C^{N_s\times T_p},
$
where \(T_p\) denotes the pilot length.
Let $\mathbf{I}\in\mathbb{R}^{C\times H\times W}$ denote the source image, where $C$, $H$, and $W$ represent the number of channels, height, and width, respectively.
At the transmitter, a pretrained Swin-Transformer-based JSCC (SwinT-JSCC) encoder is employed to extract compact semantic features from the source image.
Specifically, the encoder maps \(\mathbf I\) into a low-dimensional latent representation
\begin{equation}
\mathbf{Z}
=
f_{\theta_{\mathrm e}^{\star}}(\mathbf{I}),
\qquad
\mathbf{Z}\in\mathbb{R}^{L\times D},
\label{eq:semantic_encoder}
\end{equation}
where \(f_{\theta_{\mathrm e}^{\star}}(\cdot)\) denotes the pretrained semantic encoder with fixed parameters \(\theta_{\mathrm e}^{\star}\), \(L\) is the latent sequence length, and \(D\) is the feature dimension.
The latent representation is then quantized by a scalar quantizer $\mathcal{Q}(\cdot)$, yielding
$
\mathbf{U}
=
\mathcal{Q}(\mathbf{Z}),
\mathbf{U}\in\mathbb{R}^{L\times D}.
\label{eq:quantization}
$
To facilitate physical-layer transmission, the quantized latent $\mathbf{U}$ is mapped into a complex baseband signal through a deterministic layer mapping operator
\begin{equation}
\mathbf{X}
=
\mathcal{T}(\mathbf{U}),
\qquad
\mathcal{T}:\mathbb{R}^{L\times D}\rightarrow\mathbb{C}^{N_s\times T_d},
\label{eq:transmit_mapping}
\end{equation}
where $\mathbf{X}$ denotes the MIMO baseband signal and $T_d$ is the number of channel uses.
For clarity, the mapping $\mathcal{T}(\cdot)$ is detailed as follows.

First, $\mathbf{U}$ is vectorized into a one-dimensional sequence
\begin{equation}
\mathbf{u}
=
\mathrm{vec}(\mathbf{U})
=
[u_1,u_2,\ldots,u_{LD}]^{\mathsf T}
\in\mathbb{R}^{LD}.
% \label{eq:vectorization}
\end{equation}
To construct complex symbols, two adjacent real-valued entries are grouped into the in-phase and quadrature components.
If $LD$ is odd, one zero is appended to the tail of $\mathbf{u}$ for pairing.
Let 
$
\widetilde{\mathbf{u}}
=
[\widetilde{u}_1,\widetilde{u}_2,\ldots,\widetilde{u}_{2N_c}]^{\mathsf T}
\in\mathbb{R}^{2N_c}
$
denote the zero-padded sequence, where $N_c=\lceil LD/2\rceil$ is the total number of complex channel symbols.
The resulting complex symbol sequence is given by
\begin{equation}
r_m
=
\widetilde{u}_{2m-1}
+
j\widetilde{u}_{2m},
\qquad
m=1,2,\ldots,N_c.
\label{eq:iq_mapping}
\end{equation}

To match the spatial degrees of freedom of the MIMO channel, we reconstruct the one-dimensional symbol stream $\mathbf{r}=[r_1,r_2,\ldots,r_{N_c}]^{\mathsf T}$ into $N_s$ parallel streams.
Let $T_d=\lceil N_c/N_s\rceil$ denote the required number of channel uses.
After zero-padding, the symbol stream is rearranged into
$\mathbf{R}
=
\mathrm{reshape}(\mathbf{r},N_s,T_d)
\in\mathbb{C}^{N_s\times T_d}.
\label{eq:reshape}$
To satisfy the average transmit power constraint, $\mathbf{R}$ is normalized as
\begin{equation}
\mathbf X
=
\frac{\mathbf R}
{\sqrt{\|\mathbf R\|_{\mathrm F}^{2}/(N_sT_d)}},
\label{eq:power_norm}
\end{equation}
where the matrix $\mathbf{X}\in\mathbb{C}^{N_s\times T_d}$ is the complex baseband signal transmitted over the MIMO channel.
Before MIMO transmission, both pilot and semantic data blocks are precoded by the same QR-orthonormalized random matrix $\mathbf F\in\mathbb C^{N_t\times N_s}$ satisfying $\mathbf F^H\mathbf F=\mathbf I_{N_s}$.
Accordingly, the equivalent channel after precoding is defined as
\begin{equation}
   \mathbf H_{\mathrm e}=\mathbf H\mathbf F\in\mathbb C^{N_r\times N_s}. 
\end{equation}
This equivalent channel is shared by the pilot and semantic data blocks and will be estimated at the receiver based on the pilot observation.

\subsection{Semantic Receiver Design}
\label{subsec:receiver_design}

At the receiver, the received frame is first divided into a pilot observation block and a semantic data observation block. 
The pilot block is used to acquire the MIMO channel, whereas the semantic data block is used to recover the transmitted semantic representation. 
The corresponding received signals are given by
\begin{equation}
\mathbf Y_{\mathrm p}
=
\mathbf H_{\mathrm e}\mathbf P
+
\mathbf N_{\mathrm p},
\label{eq:pilot_model}
\end{equation}
\begin{equation}
\mathbf Y_{\mathrm d}
=
\mathbf H_{\mathrm e}\mathbf X
+
\mathbf N_{\mathrm d},
\label{eq:data_model}
\end{equation}
where $\mathbf N_{\mathrm p}$ and $\mathbf N_{\mathrm d}$ denote AWGN matrices whose entries independently follow $\mathcal{CN}(0,\sigma_n^2)$.
The receiver first obtains a coarse estimate of the channel from the pilot observation. 
The LS estimate is given by
\begin{equation}
\mathbf H_{\mathrm{LS}}
=
\mathbf Y_{\mathrm p}\mathbf P^{H}
\left(
\mathbf P\mathbf P^{H}
\right)^{\dagger}.
\label{eq:ls}
\end{equation}
This estimate provides an informative but noise-contaminated channel state. 
To mitigate the LS estimation error before semantic equalization, we introduce a channel restoration flow matching (CRFM) module, which refines the coarse estimate as
\begin{equation}
\widehat{\mathbf H}
=
\mathcal R_{\mathrm H}
\left(
\mathbf H_{\mathrm{LS}}
\right),
\label{eq:cdfm_channel_restoration}
\end{equation}
where $\mathcal R_{\mathrm H}(\cdot)$ denotes the CRFM operator\footnote{Here, all channel states fed into the CRFM module are first mapped into real-valued tensors by stacking their real and imaginary parts, i.e., $\mathbb C^{N_r\times N_s} \rightarrow \mathbb R^{2\times N_r\times N_s}$.}.
Based on the refined channel, the receiver constructs the minimum mean square error (MMSE) equalizer as
\begin{equation}
\mathbf W
\left(
\widehat{\mathbf H}
\right)
=
\left(
\widehat{\mathbf H}^{H}
\widehat{\mathbf H}
+
\sigma_n^2\mathbf I_{N_s}
\right)^{-1}
\widehat{\mathbf H}^{H}.
\label{eq:mmse_equalizer_refined}
\end{equation}
The equalized semantic signal is then obtained by
\begin{equation}
\widehat{\mathbf X}_{\mathrm{MMSE}}
=
\mathbf W
\left(
\widehat{\mathbf H}
\right)
\mathbf Y_{\mathrm d}.
\label{eq:coarse_equalized_signal}
\end{equation}
Through the deterministic receive-side demapping operator $\mathcal T^{-1}(\cdot)$, the corresponding coarse semantic latent representation is obtained as
$
\widehat{\mathbf U}_{\mathrm{MMSE}}
=
\mathcal T^{-1}
\left(
\widehat{\mathbf X}_{\mathrm{MMSE}}
\right).
\label{eq:coarse_semantic_latent}
$
Although channel restoration and MMSE equalization suppress part of the physical-layer distortion, $\widehat{\mathbf U}_{\mathrm{MMSE}}$ may still contain residual perturbations caused by imperfect-CSI, equalization mismatch, and data-domain noise. 
Therefore, a semantic restoration flow matching (SRFM) module is further employed to correct the coarse semantic latent representation
\begin{equation}
\widehat{\mathbf U}
=
\mathcal R_{\mathrm U}
\left(
\widehat{\mathbf U}_{\mathrm{MMSE}}
\right),
\label{eq:sdfm_semantic_restoration}
\end{equation}
where $\mathcal R_{\mathrm U}(\cdot)$ denotes the SRFM operator.
Finally, the restored semantic latent representation is fed into the pretrained SwinT-JSCC decoder to reconstruct the source image
\begin{equation}
\widehat{\mathbf I}
=
g_{\theta_{\mathrm d}^{\star}}
\left(
\widehat{\mathbf U}
\right),
\label{eq:semantic_decoder_reconstruction}
\end{equation}
where \(g_{\theta_{\mathrm d}^{\star}}(\cdot)\) denotes the pretrained semantic decoder with fixed parameters \(\theta_{\mathrm d}^{\star}\).
\subsection{Problem Formulation}
\label{subsec:problem_formulation}

In imperfect-CSI-MIMO semantic communication, semantic recovery at the receiver is essentially performed in a cascaded manner, since the semantic equalizer is constructed based on channel estimates rather than the actual channel. 
Consequently, channel estimation errors are not confined to the channel domain but are converted by the equalizer into structured distortions in the transmitted semantic signal, thereby affecting the latent semantic representation.

Substituting \eqref{eq:pilot_model} into \eqref{eq:ls}, we obtain
\begin{equation}
\mathbf H_{\mathrm{LS}} = \mathbf H_{\mathrm e} + \underbrace{ \mathbf N_{\mathrm p}\mathbf P^{\mathrm H}\left( \mathbf P\mathbf P^{\mathrm H} \right)^{\dagger} }_{\Delta\mathbf H},
\label{eq:ls_channel_error_model_pf}
\end{equation}
where $\Delta\mathbf H$ denotes the  channel estimation error.
Based on this imperfect-CSI, the MMSE equalizer given by \eqref{eq:mmse_equalizer_refined} is
\begin{equation}
\mathbf W\!\left(\mathbf H_{\mathrm{LS}}\right)
=
\left[
\left(\mathbf H_{\mathrm{LS}}\right)^{H}
\mathbf H_{\mathrm{LS}}
+
\sigma_n^2\mathbf I_{N_s}
\right]^{-1}
\left(\mathbf H_{\mathrm{LS}}\right)^{H}.
\label{eq:ls_based_mmse_pf}
\end{equation}

For the semantic data block, substituting \eqref{eq:mmse_equalizer_refined} into the MMSE equalization gives
\begin{equation}
\begin{aligned}
\mathbf X_{\mathrm{MMSE}}
&=
\mathbf W\!\left(\mathbf H_{\mathrm{LS}}\right)\mathbf Y_{\mathrm d}
\\
&=
\mathbf W\!\left(\mathbf H_{\mathrm e}+\Delta\mathbf H\right)
\left(
\mathbf H_{\mathrm e}\mathbf X+\mathbf N_{\mathrm d}
\right)
\\
&=
\mathbf W\!\left(\mathbf H_{\mathrm e}+\Delta\mathbf H\right)
\mathbf H_{\mathrm e}\mathbf X
+
\mathbf W\!\left(\mathbf H_{\mathrm e}+\Delta\mathbf H\right)
\mathbf N_{\mathrm d}.
\end{aligned}
\label{eq:coarse_equalization_pf}
\end{equation}
Accordingly, the equalization error can be decomposed as
\begin{equation}
\begin{aligned}
\mathbf X_{\mathrm{MMSE}}-\mathbf X
&=
\left[
\mathbf W\!\left(\mathbf H_{\mathrm e}+\Delta\mathbf H\right)
\mathbf H_{\mathrm e}
-
\mathbf I_{N_s}
\right]\mathbf X
\\
&\quad+
\mathbf W\!\left(\mathbf H_{\mathrm e}+\Delta\mathbf H\right)
\mathbf N_{\mathrm d}.
\end{aligned}
\label{eq:equalization_error_decomposition_pf}
\end{equation}
Therefore, the channel estimation error $\lvert \Delta \mathbf H \rvert$ typically amplifies the residual equalization mismatch in $\mathbf X_{\mathrm{MMSE}}$ and further propagates it into the semantic latent representations \cite{Y_MIMO_detection}.
The above error propagation relationship indicates that it is necessary to improve the accuracy of channel estimation before performing semantic equalization. 
Therefore, we first introduce the CRFM operator $\mathcal R_{\mathrm H}(\cdot)$
\begin{equation}
\mathcal R_{\mathrm H}^{\star}
=
\arg\min_{\mathcal R_{\mathrm H}}
\mathbb E
\left[
\left\|
\mathcal R_{\mathrm H}
\left(
\mathbf H_{\mathrm{LS}}
\right)
-
\mathbf H_{\mathrm e}
\right\|_{\mathrm F}^{2}
\right].  
\label{eq:CRFM_operate}
\end{equation}
Although CRFM improves the accuracy of channel estimation, as derived in \eqref{eq:coarse_equalized_signal}, the equalized semantic latent remain a disturbed representation \cite{Wu_CDDM_2024, Xu_SPEDNSC_2025}, as follows
\begin{equation}
\widehat{\mathbf U}_{\mathrm{MMSE}}
=
\mathcal T^{-1}
\left(
\widehat{\mathbf X}_{\mathrm{MMSE}}
\right)
=
\mathbf U+\Delta\mathbf U,
\label{eq:MMSE_U}
\end{equation}
where \(\Delta\mathbf U\) denotes the effective semantic latent perturbation.
This motivates the introduction of the SRFM operator  $\mathcal R_{\mathrm U}(\cdot)$
\begin{equation}
\mathcal R_{\mathrm U}^{\star}
=
\arg\min_{\mathcal R_{\mathrm U}}
\mathbb E
\left[
\left\|
\mathcal R_{\mathrm U}
\left(
\widehat{\mathbf U}_{\mathrm{MMSE}}
\right)
-
\mathbf U
\right\|_{\mathrm F}^{2}
\right]. 
\label{eq:SRFM_operate}
\end{equation}
In short, \eqref{eq:ls_channel_error_model_pf}–\eqref{eq:equalization_error_decomposition_pf} show that channel estimation errors are converted by a CSI-based equalizer into structured equalization mismatches, which in turn induce the semantic latent disturbances described in \eqref{eq:MMSE_U}.
Although the two inverse problems of channel estimation and equalization differ in terms of signal-domain distortion and distortion characteristics, making it difficult to address them adequately with a single model, they share the same restoration structure.
Whether for channel refinement or equalization correction, their objective is to recover a clean target from observation-dependent corrupted states.
Section \ref{III} addresses this issue using a cascaded RFM framework.

\section{Cascaded Restoration Flow Matching within the SwinT-JSCC Framework}
\label{III}
Tailored for the SwinT-JSCC-based MIMO semantic communication framework, this section develops a cascaded restoration flow matching receiver for the two successive inverse recovery tasks at the receiver, namely channel restoration flow matching (CRFM) and semantic restoration flow matching (SRFM).

\subsection{Semantic Representation Learning Based on SwinT-JSCC}
As shown in Fig. \ref{Architecture}, before developing the restoration flow modules, we first pretrain the SwinT-JSCC encoder-decoder to establish a stable semantic latent representation space \cite{Liu_Swin_2021}. 
To improve perceptual realism, an adversarial loss is incorporated through a multi-scale discriminator $D$, $\widetilde{\mathbf{I}}$ denotes the interpolated sample between the real image and the reconstructed image. 
We design the loss function as follows
\begin{equation}
\mathcal{L}_{\mathrm{JSCC}}
=
\lambda_{\mathrm{mse}}\mathcal{L}_{\mathrm{mse}}
+
\lambda_{\mathrm{ssim}}\mathcal{L}_{\mathrm{ssim}}
+
\lambda_{\mathrm{adv}}\mathcal{L}_{\mathrm{adv}}
+
\lambda_{\mathrm{gp}}\mathcal{L}_{\mathrm{gp}},
\label{eq:jscc_loss_total}
\end{equation}

where
\begin{equation}
\mathcal{L}_{\mathrm{mse}}
=
\mathbb{E}\!\left[
\|\widehat{\mathbf{I}}-\mathbf{I}\|_2^2
\right],
\label{eq:mse_loss}
\end{equation}
\begin{equation}
\mathcal{L}_{\mathrm{ssim}}
=
1-\mathrm{SSIM}(\widehat{\mathbf{I}},\mathbf{I}),
\label{eq:ssim_loss}
\end{equation}
\begin{equation}
\mathcal{L}_{\mathrm{adv}}
=
-\mathbb{E}\!\left[D(\widehat{\mathbf{I}})\right],
\label{eq:adv_loss}
\end{equation}
\begin{equation}
\mathcal{L}_{\mathrm{gp}}
=
\mathbb{E}_{\widetilde{\mathbf{I}}}
\left[
\left(\|\nabla_{\widetilde{\mathbf{I}}}D(\widetilde{\mathbf{I}})\|_2-1\right)^2
\right],
\label{eq:gp_loss}
\end{equation}
where $\lambda_{\mathrm{mse}}$, $\lambda_{\mathrm{ssim}}$, $\lambda_{\mathrm{adv}}$, and $\lambda_{\mathrm{gp}}$ are nonnegative weighting coefficients.
By minimizing \(\mathcal{L}_{\mathrm{JSCC}}\), the pretrained semantic encoder and decoder are obtained as
\begin{equation}
(\theta_{\mathrm e}^{\star},\theta_{\mathrm d}^{\star})
=
\arg\min_{\theta_{\mathrm e},\theta_{\mathrm d}}
\mathcal{L}_{\mathrm{JSCC}},
\label{eq:jscc_pretraining_optimization}
\end{equation}
which yield the semantic encoding and decoding mappings
\(f_{\theta_{\mathrm e}^{\star}}(\cdot)\) and \(g_{\theta_{\mathrm d}^{\star}}(\cdot)\), respectively.
After pretraining, these two mappings are fixed to provide a stable latent representation space.
\subsection{Unified Flow Matching Framework}
In this subsection, we first introduce a unified flow matching framework for the two recovery tasks described in \eqref{eq:CRFM_operate} and \eqref{eq:SRFM_operate}.
To simplify the derivation, we use \((\mathbf S^0,\mathbf S^1)\) to denote a generic source-target pair along a restoration path. Specifically,
\begin{equation}
(\mathbf S^0,\mathbf S^1)
=
\begin{cases}
(\mathbf H^0,\mathbf H^1), & \text{for the CRFM operator},\\[0.3em]
(\mathbf U^0,\mathbf U^1), & \text{for the SRFM operator}.
\end{cases}
\label{eq:unified_state_pair}
\end{equation}
where 
\(\mathbf H^0,\mathbf H^1\in\mathbb R^{2\times N_r\times N_s}\) 
denote the source and target endpoints in the real-valued channel tensor space, respectively, and 
\(\mathbf U^0,\mathbf U^1\in\mathbb R^{L\times D}\) 
denote the source and target endpoints in the semantic latent space, respectively. 
FM learns a time-dependent vector field $\mathbf{v}_t^\star(\mathbf{S})$ that transports a source distribution $p_0$ to a target distribution $p_1$ \cite{lipman2023flow}. 
Let $\{p_t\}_{t\in[0,1]}$ be a probability path connecting $p_0$ and $p_1$. 
The corresponding marginal velocity field $\mathbf{v}_t^\star(\mathbf{S})$ is defined through the continuity equation
\begin{equation}
\frac{\partial p_t(\mathbf{S})}{\partial t}
+
\nabla_{\mathbf{S}}\!\cdot\!\bigl[p_t(\mathbf{S})\,\mathbf{v}_t^\star(\mathbf{S})\bigr] = 0.
\label{eq:fm_continuity_revised}
\end{equation}
Ideally, one would learn a neural field $\mathbf{v}_{\theta}(\mathbf{S}_{t},t)$ by minimizing
\begin{equation}
\mathcal{L}_{\mathrm{FM}}(\theta)
=
\int_0^1
\mathbb{E}_{\mathbf{S}_t\sim p_t}
\!\left[
\left\|
\mathbf{v}_{\theta}(\mathbf{S}_t,t)
-
\mathbf{v}_t^\star(\mathbf{S})
\right\|_2^2
\right]\mathrm{d}t.
\label{eq:marginal_fm_loss_revised}
\end{equation}
However, directly minimizing \eqref{eq:marginal_fm_loss_revised} is generally intractable since the marginal velocity field is unavailable. According to conditional flow matching (CFM), this objective can be replaced by a tractable conditional velocity regression with the same optimal marginal vector field \cite{lipman2023flow}, i.e., 
\begin{equation} 
\mathcal L_{\mathrm{CFM}}(\theta) = \mathbb E_{t,\,p_t(\mathbf S_t|\mathbf S^1)} \left[ \left\| \mathbf v_{\theta}(\mathbf S_t,t) - \mathbf v_t^\star(\mathbf S_t|\mathbf S^1) \right\|_2^2 \right]. \label{eq:cfm_objective} 
\end{equation} 
For the optimal transport path \cite{kerrigan2024dynamic} 
\begin{equation} 
\mathbf S_t=t\mathbf S^1+(1-t)\mathbf S^0, \label{eq:linear_conditional_path} 
\end{equation} 
In standard FM, the two endpoints of the probability path are given by
$\mathbf S^0 \sim p_0$ and $\mathbf S^1 \sim p_1$, where \(\mathbf S^0\sim p_0\) denotes the random source state, which is usually chosen as a standard Gaussian noise, i.e., $\mathbf p_0 {\sim} \mathcal N(0,1)$.
The resulting conditional velocity field is as follows
\begin{equation} 
\mathbf v_t^\star(\mathbf S_t|\mathbf S^1) = \frac{\mathrm d\mathbf S_t}{\mathrm dt} = \mathbf S^1-\mathbf S^0. \label{eq:conditional_velocity_closed_form} 
\end{equation}
Therefore, the CFM loss is rewritten as 
\begin{equation} 
\mathcal L_{\mathrm{CFM}}(\theta) = \mathbb E_{t,\,(\mathbf S^0,\mathbf S^1)} \left[ \left\| \mathbf v_{\theta}(\mathbf S_t,t) - (\mathbf S^1-\mathbf S^0) \right\|_2^2 \right].
\label{eq:cfm_loss_final} 
\end{equation}

Based on the probabilistic path model described above, the generation process can be viewed as a transition from Gaussian noise to the target data distribution.
However, for communication recovery tasks, this noise-to-data paradigm is not always the most natural choice.
Unlike unconditional generation, the receiver’s goal is to recover the specific channel or semantic latent state corresponding to the currently received observations. 
Therefore, the recovered results must not only be distribution-wise reasonable but also consistent with the received pilot signals or semantic observations.
To address this issue, we propose training and inference schemes for the RFM model in Sections \ref{subsec:Train of Restoration Flow Matching model} and \ref{subsec:ode_based_inference}, respectively.

\subsection{Training of Restoration Flow Matching Models}
\label{subsec:Train of Restoration Flow Matching model}
% \begin{figure}[!t]
% \centering
% \includegraphics[width=3.2in]{UNet.png}
% \caption{Impact of different channel-estimation algorithms on semantic reconstruction in terms of MS-SSIM. } 
% \label{fig:msssim_diff_CE_algorithm}
% \end{figure}

In this subsection, we train the CRFM and SRFM modules as coarse-to-clean restoration flow models. 
For each clean target, we construct a controlled perturbed source state and learn a constant-velocity path toward the paired clean state.
The training of the CRFM velocity field is parameterized using a UNet-based network \cite{Ho_DDPM_2020}, while the training of the SRFM velocity field is parameterized using a DiT-based network \cite{Scalable_2023}.

Specifically, we inject controlled perturbation into the clean channel state and the clean semantic latent variables as the source states, while using the corresponding clean states as the recovery targets, as follows
\begin{equation}
\mathbf{S}_{\tau}^{0}
=
\mathbf{S}^1
+
\tau\boldsymbol{\xi},
\qquad
\boldsymbol{\xi}\sim\mathcal{N}(\mathbf{0},\mathbf I),
\label{eq:ogdfm_init}
\end{equation}
where $\tau>0$ controls the perturbation strength around the clean-state manifold.
Here, $\mathbf S_{\tau}^{0}$ represents the perturbed channel state in CRFM, or the perturbed semantic latent state in SRFM.
In this way, the source states are locally perturbed states around the clean state manifold. 
This provides a manageable training path from coarse to clean states for learning the restoration vector field used for channel refinement and equalization correction.

For $t\in[0,1]$, the associated interpolation path is defined as
\begin{equation}
\mathbf{S}_{\tau}^t
=
t\mathbf{S}^1
+
(1-t)\mathbf{S}_{\tau}^{0}
=
\mathbf{S}^1
+
(1-t)\tau\boldsymbol{\xi}.
\label{eq:ogdfm_path}
\end{equation}
Taking the derivative of \eqref{eq:ogdfm_path} with respect to $t$ gives the conditional velocity
\begin{equation}
\mathbf{u}_{\tau}^{t}
\triangleq
\frac{\mathrm d \mathbf{S}_{\tau}^{t}}{\mathrm dt}
=
-\tau\boldsymbol{\xi}.
\label{eq:ogdfm_conditional_velocity}
\end{equation}
This path starts from the perturbed source state $\mathbf S_\tau^0$ at $t=0$ and reaches the clean target state $\mathbf S^1$ at $t=1$. 
Therefore, the restoration flow field can be trained through the following single-anchor velocity regression objective
\begin{equation}
\mathcal{L}_{\mathrm{RFM}}(\theta;\tau)
=
\mathbb{E}_{\mathbf S^1,\boldsymbol{\xi},t}
\left[
\left\|
\mathbf v_{\theta}(\mathbf S_{\tau}^{t},t)
-
\mathbf u_{\tau}^{t}
\right\|_2^2
\right].
\label{eq:rfm_single_anchor_loss}
\end{equation}

Accordingly, the generic restoration path in \eqref{eq:ogdfm_path} is instantiated for CRFM and SRFM as
\begin{align}
\mathbf H_{\tau}^{t}
&=
\mathbf H^{1}
+
(1-t)\tau\boldsymbol{\xi},
\label{eq:channel_restoration_path}
\\
\mathbf U_{\tau}^{t}
&=
\mathbf U^{1}
+
(1-t)\tau\boldsymbol{\xi},
\label{eq:semantic_restoration_path}
\end{align}
where $\mathbf H_{\tau}^{t}$ and $\mathbf U_{\tau}^{t}$ denote the intermediate restoration states at time $t$ in the channel and semantic latent domains, respectively. For notational simplicity, $\boldsymbol{\xi}$ denotes an independently sampled Gaussian perturbation in each domain, with dimensions conformable to $\mathbf H^{1}$ or $\mathbf U^{1}$.

Since the two restoration paths are linear in $t$, their target velocities are both given by $-\tau\boldsymbol{\xi}$.
Accordingly, the CRFM and SRFM velocity fields are learned in their respective domains by minimizing the following task-specific velocity-regression objectives
\begin{align}
\mathcal{L}_{\mathrm{H}}(\theta_{\mathrm H};\tau)
&=
\mathbb{E}_{\mathbf H^{1},\boldsymbol{\xi},t}
\left[
\left|
\mathbf v_{\theta_{\mathrm H}}(\mathbf H_{\tau}^{t},t)
+
\tau\boldsymbol{\xi}
\right|_2^2
\right],
\label{eq:crfm_loss_single_anchor} \\
\mathcal{L}_{\mathrm{U}}(\theta_{\mathrm U};\tau)
&=
\mathbb{E}_{\mathbf U^{1},\boldsymbol{\xi},t}
\left[
\left|
\mathbf v_{\theta_{\mathrm U}}(\mathbf U_{\tau}^{t},t)
+
\tau\boldsymbol{\xi}
\right|_2^2
\right],
\label{eq:srfm_loss_single_anchor}
\end{align}
where $\theta_{\mathrm H}$ and $\theta_{\mathrm U}$ are the trainable parameters of the CRFM and SRFM velocity networks, respectively.
These objectives force the predicted velocity fields to match the constant restoration directions along the channel-domain and semantic-domain paths. Therefore, the learned networks are encouraged to progressively transport perturbed states toward their corresponding clean targets.
The following proposition shows that the resulting optimal restoration velocity is proportional to the score of the Gaussian-smoothed clean-state distribution.

\emph{Proposition 1:}
\textit{Under the fixed-perturbation path in \eqref{eq:ogdfm_path} with the standard Gaussian perturbation in \eqref{eq:ogdfm_init}, let
\(\mathbf v_\tau^\star(\mathbf s,t)\) be the population-risk minimizer of \eqref{eq:rfm_single_anchor_loss} over square-integrable velocity fields, where \(\tau>0\) and \(t\in[0,1)\).
Assume that the marginal density \(p_\tau^t(\mathbf s)\) of \(\mathbf S_\tau^t\) is differentiable and positive.
Then,}
\begin{equation}
\mathbf v_\tau^\star(\mathbf s,t)
=
(1-t)\tau^2
\nabla_{\mathbf s}
\log p_\tau^t(\mathbf s).
\label{eq:dfm_score_velocity}
\end{equation}

\noindent\emph{Proof:}
For a fixed \(t\in[0,1)\), \eqref{eq:rfm_single_anchor_loss} reduces to a squared-error regression problem that maps the intermediate state \(\mathbf S_\tau^t\) to the target velocity \(\mathbf u_\tau^t\) \cite{lipman2023flow}. 
Therefore, the population-risk minimizer is given by the conditional mean estimator
\begin{equation}
\mathbf v_\tau^\star(\mathbf s,t)
=
\mathbb E
\left[
\mathbf u_\tau^t
\;\middle|\;
\mathbf S_\tau^t=\mathbf s
\right].
\label{eq:proof_conditional_velocity}
\end{equation}
From the restoration path in \eqref{eq:ogdfm_path} and the target velocity in
\eqref{eq:ogdfm_conditional_velocity}, we have
\begin{equation}
\mathbf u_\tau^t
=
-\tau\boldsymbol{\xi}
=
\frac{\mathbf S^1-\mathbf S_\tau^t}{1-t}.
\label{eq:proof_velocity_relation}
\end{equation}
Substituting \eqref{eq:proof_velocity_relation} into
\eqref{eq:proof_conditional_velocity} gives
\begin{equation}
\mathbf v_\tau^\star(\mathbf s,t)
=
\frac{
\mathbb E
\left[
\mathbf S^1
\;\middle|\;
\mathbf S_\tau^t=\mathbf s
\right]
-
\mathbf s
}{1-t}.
\label{eq:proof_posterior_mean_velocity}
\end{equation}

Under the Gaussian perturbation model in \eqref{eq:ogdfm_init},
\(\mathbf S_\tau^t\mid \mathbf S^1\sim
\mathcal N(\mathbf S^1,\sigma_{\tau,t}^2\mathbf I)\), where
$
\sigma_{\tau,t}^2=(1-t)^2\tau^2 .
$
Therefore, \(p_\tau^t(\mathbf s)\) is the Gaussian-smoothed density of the clean state. 
By Tweedie's identity \cite{Kim2021Noise2Score}, we have
\begin{equation}
\mathbb E
\left[
\mathbf S^1
\;\middle|\;
\mathbf S_\tau^t=\mathbf s
\right]
-
\mathbf s
=
\sigma_{\tau,t}^2
\nabla_{\mathbf s}\log p_\tau^t(\mathbf s).
\label{eq:proof_tweedie_identity}
\end{equation}
Substituting \eqref{eq:proof_tweedie_identity} into
\eqref{eq:proof_posterior_mean_velocity} yields
\begin{equation}
\mathbf v_\tau^\star(\mathbf s,t)
=
\frac{\sigma_{\tau,t}^2}{1-t}
\nabla_{\mathbf s}\log p_\tau^t(\mathbf s)
=
(1-t)\tau^2
\nabla_{\mathbf s}\log p_\tau^t(\mathbf s).
\end{equation}
This completes the proof.
\hfill $\blacksquare$

\noindent\emph{Remark 1:}
Proposition 1 provides a score-based interpretation of the fixed-perturbation RFM objective.
Specifically, the population-optimal velocity field does not merely fit the artificially injected Gaussian perturbation of each training sample. 
Rather, it captures a local restoration direction characterized by the score of the Gaussian-smoothed clean-state density, which guides perturbed states toward the clean-state manifold.
These directions point toward regions of high probability on the “clean channel” or the “clean semantic latent”.
Hence, the online initial state need not exactly follow the artificial Gaussian perturbation model used in training, as long as the receiver-side coarse estimate lies within the clean-state neighborhood.

However, a single perturbation anchor is insufficient to characterize the heterogeneous distortion conditions encountered by practical receivers \cite{song2021score, song2019generative}.
Specifically, a small perturbation anchor generates training states close to the clean-state manifold and is suitable for fine-grained refinement, but it provides limited coverage for severely corrupted coarse receiver states. 
In contrast, a large perturbation anchor enlarges the coverage of heavily distorted states, but may weaken the local restoration accuracy around the clean-state manifold. 

To accommodate different distortion levels, we introduce a dual-anchor perturbation training strategy with two perturbation anchors \(0<\tau_{\ell}<\tau_{h}\).
The low perturbation anchor \(\tau_{\ell}\) focuses on local refinement around the clean-state manifold, whereas the high perturbation anchor \(\tau_{h}\) improves the coverage of strongly corrupted receiver states.
The sensitivity to the perturbation-anchor selection is empirically evaluated in Section~\ref{subsec:rfm_sensitivity}.

Specifically, the low- and high-anchor restoration paths are defined as
\begin{align}
\mathbf S_{\tau_\ell}^{t}
&=
\mathbf S^{1}
+
(1-t)\tau_{\ell}\boldsymbol{\xi},
&
\mathbf S_{\tau_h}^{t}
&=
\mathbf S^{1}
+
(1-t)\tau_{h}\boldsymbol{\xi},
\label{eq:dual_perturbation_paths}
\end{align}
The corresponding conditional restoration velocities are defined as
\begin{align}
\mathbf u_{\tau_\ell}^{t}
&\triangleq
\frac{\mathrm d\mathbf S_{\tau_\ell}^{t}}{\mathrm dt}
=
-\tau_{\ell}\boldsymbol{\xi},
&
\mathbf u_{\tau_h}^{t}
&\triangleq
\frac{\mathrm d\mathbf S_{\tau_h}^{t}}{\mathrm dt}
=
-\tau_{h}\boldsymbol{\xi}.
\label{eq:dual_perturbation_velocities}
\end{align}

\textbf{Dual-anchor Perturbation Training Objective:}
Based on \eqref{eq:dual_perturbation_paths} and \eqref{eq:dual_perturbation_velocities}, the dual-perturbation restoration flow matching objective is formulated as
\begin{equation}
\begin{aligned}
\mathcal{L}_{\mathrm{RFM}}(\theta;\tau_{\ell},\tau_{h})
&=
\omega_{\ell}
\mathbb E_{\mathbf S^1,\boldsymbol{\xi},t}
\left[
\left\|
\mathbf v_{\theta}
\left(
\mathbf S_{\tau_\ell}^{t},t
\right)
-
\mathbf u_{\tau_\ell}^{t}
\right\|_2^2
\right]
\\
&\quad+
\omega_{h}
\mathbb E_{\mathbf S^1,\boldsymbol{\xi},t}
\left[
\left\|
\mathbf v_{\theta}
\left(
\mathbf S_{\tau_h}^{t},t
\right)
-
\mathbf u_{\tau_h}^{t}
\right\|_2^2
\right],
\end{aligned}
\label{eq:dual_perturbation_training_loss}
\end{equation}
In this work, we set $\omega_\ell=\omega_h=1/2$ to balance near-manifold refinement and large-error correction.
Based on the single-anchor restoration objectives in 
\eqref{eq:crfm_loss_single_anchor} and \eqref{eq:srfm_loss_single_anchor}, 
the proposed dual-anchor perturbation training objectives for CRFM and SRFM are obtained by weighted aggregation over the low- and high-perturbation anchors
\begin{align}
\theta_{\mathrm H}^{\star}
&=
\arg\min_{\theta_{\mathrm H}}
\underbrace{
\left\{
\omega_{\ell}
\mathcal L_{\mathrm H}(\theta_{\mathrm H};\tau_{\ell})
+
\omega_{h}
\mathcal L_{\mathrm H}(\theta_{\mathrm H};\tau_{h})
\right\}
}_{\triangleq\,
\mathcal L_{\mathrm H}(\theta_{\mathrm H};\tau_{\ell},\tau_{h})},
\label{eq:crfm_dual_anchor_loss}
\\
\theta_{\mathrm U}^{\star}
&=
\arg\min_{\theta_{\mathrm U}}
\underbrace{
\left\{
\omega_{\ell}
\mathcal L_{\mathrm U}(\theta_{\mathrm U};\tau_{\ell})
+
\omega_{h}
\mathcal L_{\mathrm U}(\theta_{\mathrm U};\tau_{h})
\right\}
}_{\triangleq\,
\mathcal L_{\mathrm U}(\theta_{\mathrm U};\tau_{\ell},\tau_{h})},
\label{eq:srfm_dual_anchor_loss}
\end{align}
where \(\theta_{\mathrm H}^{\star}\) and \(\theta_{\mathrm U}^{\star}\) are the optimized parameters for CRFM and SRFM, respectively.
It is worth noting that each of these two modules learns a mixed restoration vector field on the composite trajectory distribution induced by these two perturbation anchors.
\begin{algorithm}[!t]
\caption{Training of CRFM}
\label{alg:crfm_training}
\begin{algorithmic}[1]
\Require Clean channel tensor dataset $\mathcal D_{\mathrm H}$, perturbation anchors
$\{\tau_\ell,\tau_h\}$, batch size $B$, iterations $N_{\mathrm{it}}$.
\Ensure Trained channel velocity field $\mathbf v_{\theta_{\mathrm H}^{\star}}(\cdot,t)$.
\For{$n=1,\ldots,N_{\mathrm{it}}$}
    \State Sample clean channel targets
    $\{\mathbf H^{1,(b)}\}_{b=1}^{B}$ from $\mathcal D_{\mathrm H}$.
    \State Sample $t^{(b)}\sim\mathcal U(0,1)$ and
    $\tau^{(b)}\in\{\tau_\ell,\tau_h\}$.
    \State Construct the channel restoration state and target velocity
    according to \eqref{eq:dual_perturbation_paths} and
    \eqref{eq:dual_perturbation_velocities}.
    \State Update $\theta_{\mathrm H}$ by gradient step on
    $\mathcal L_{\mathrm H}(\theta_{\mathrm H};\tau_\ell,\tau_h)$
    in \eqref{eq:crfm_dual_anchor_loss}.
\EndFor
\end{algorithmic}
\end{algorithm}

\begin{algorithm}[!t]
\caption{Training of SRFM}
\label{alg:srfm_training}
\begin{algorithmic}[1]
\Require Image dataset $\mathcal D_{\mathrm I}$, pretrained semantic encoder
$f_{\theta_{\mathrm e}}(\cdot)$, quantizer $\mathcal Q(\cdot)$, perturbation anchors
$\{\tau_\ell,\tau_h\}$, batch size $B$, iterations $N_{\mathrm{it}}$.
\Ensure Trained semantic velocity field $\mathbf v_{\theta_{\mathrm U}^{\star}}(\cdot,t)$.
\For{$n=1,\ldots,N_{\mathrm{it}}$}
    \State Sample images $\{\mathbf I^{(b)}\}_{b=1}^{B}$ from $\mathcal D_{\mathrm I}$ and obtain
    clean semantic targets
    $\mathbf U^{1,(b)}=\mathcal Q(f_{\theta_{\mathrm e}}(\mathbf I^{(b)}))$.
    \State Sample $t^{(b)}\sim\mathcal U(0,1)$ and
    $\tau^{(b)}\in\{\tau_\ell,\tau_h\}$.
    \State Construct the semantic restoration state and target velocity
    according to \eqref{eq:dual_perturbation_paths} and
    \eqref{eq:dual_perturbation_velocities}.
    \State Update $\theta_{\mathrm U}$ by gradient step on
    $\mathcal L_{\mathrm U}(\theta_{\mathrm U};\tau_\ell,\tau_h)$
    in \eqref{eq:srfm_dual_anchor_loss}.
\EndFor
\end{algorithmic}
\end{algorithm}
\begin{figure*}[!t]
\centering
\includegraphics[width=5in]{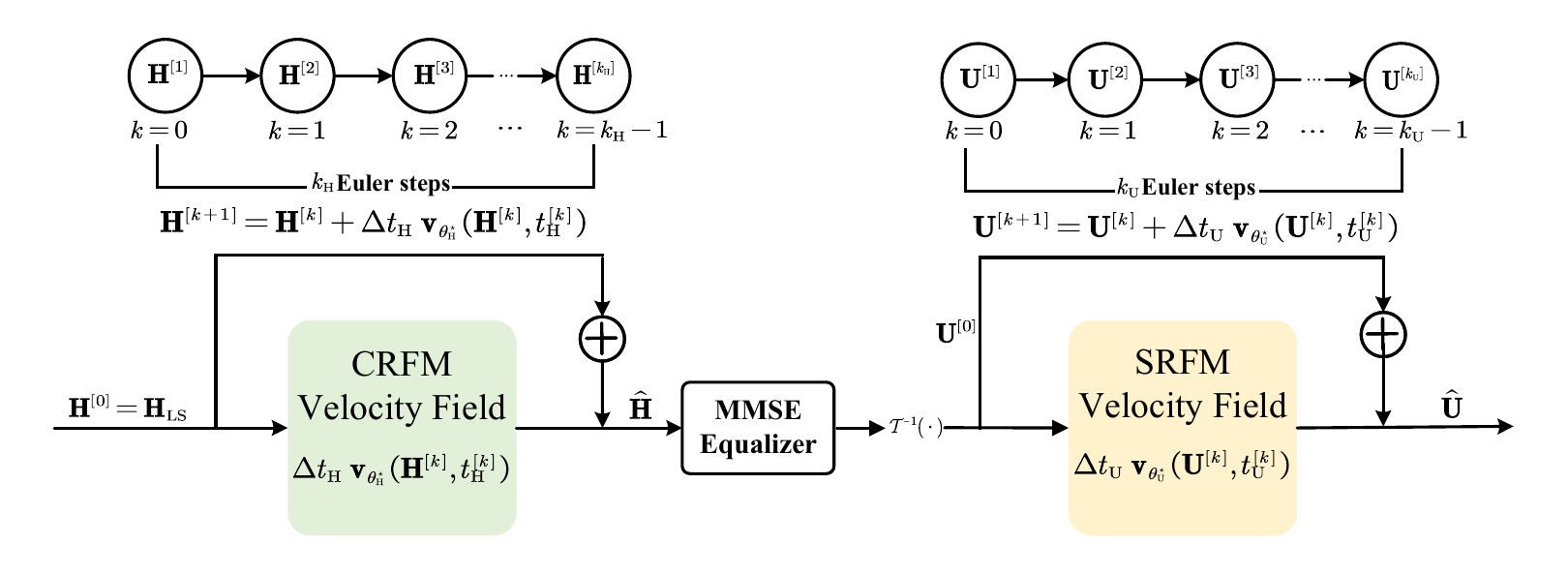}
\caption{Overall workflow of the proposed cascaded RFM inference.}
\label{fig:algorithm_flow}
\end{figure*}
The following analysis shows that the dual-perturbation objectives in \eqref{eq:crfm_dual_anchor_loss} and \eqref{eq:srfm_dual_anchor_loss} can be interpreted as a velocity regression problem over a mixture of restoration-path distributions. This interpretation explains how a single shared velocity field can adapt to different receiver-side corruption levels without requiring a densely sampled perturbation schedule.

Let $\mu_{\tau_\ell}^{t}$ and $\mu_{\tau_h}^{t}$ denote the joint distributions of
$(\mathbf S_{\tau_\ell}^{t},\mathbf u_{\tau_\ell}^{t})$ and
$(\mathbf S_{\tau_h}^{t},\mathbf u_{\tau_h}^{t})$, respectively.
The mixture training law is given by
\begin{equation}
\mu_{\tau_\ell,\tau_h}^{t}
=
\omega_\ell \mu_{\tau_\ell}^{t}
+
\omega_h \mu_{\tau_h}^{t}.
\label{eq:dual_perturbation_joint_law}
\end{equation}

Let $p_{\tau}^{t}(\mathbf s)$ denote the marginal density of
$\mathbf S_{\tau}^{t}$ for $\tau\in\{\tau_\ell,\tau_h\}$.
Since $\mathbf S_{\tau}^{t}=\mathbf S^1+(1-t)\tau\boldsymbol{\xi}$,
we have
\begin{equation}
p_{\tau}^{t}(\mathbf s)
=
\mathbb E_{\mathbf S^1\sim p_1}
\left[
\mathcal N
\left(
\mathbf s;\mathbf S^1,
(1-t)^2\tau^2\mathbf I
\right)
\right],
\quad
\tau\in\{\tau_\ell,\tau_h\}.
\label{eq:anchor_marginal_density_compact}
\end{equation}
Accordingly,
\begin{equation}
p_{\tau_\ell,\tau_h}^{t}(\mathbf s)
=
\omega_\ell p_{\tau_\ell}^{t}(\mathbf s)
+
\omega_h p_{\tau_h}^{t}(\mathbf s).
\label{eq:dual_perturbation_marginal_density}
\end{equation}

For each anchor, define the corresponding population optimal velocity field as
\begin{equation}
\mathbf v_{\tau}^{\star}(\mathbf s,t)
=
\mathbb E_{\mu_{\tau}^{t}}
\left[
\mathbf u
\,\middle|\,
\mathbf S=\mathbf s
\right],
\quad
\tau\in\{\tau_\ell,\tau_h\}.
\label{eq:anchor_velocity_field_compact}
\end{equation}

\emph{Proposition 2:}
\textit{For a fixed $t\in[0,1)$, consider the population squared-risk minimizer under the mixture law $\mu_{\tau_\ell,\tau_h}^{t}$,
\begin{equation}
\mathbf v_{\tau_\ell,\tau_h}^{\star}(\cdot,t)
=
\arg\min_{\mathbf v}
\mathbb E_{\mu_{\tau_\ell,\tau_h}^{t}}
\left[
\left\|
\mathbf v(\mathbf S,t)-\mathbf u
\right\|_2^2
\right].
\label{eq:dual_perturbation_population_regression}
\end{equation}
For any $\mathbf s$ with $p_{\tau_\ell,\tau_h}^{t}(\mathbf s)>0$, the optimal velocity field satisfies
\begin{equation}
\mathbf v_{\tau_\ell,\tau_h}^{\star}(\mathbf s,t)
=
\gamma_{\tau_\ell}(\mathbf s,t)
\mathbf v_{\tau_\ell}^{\star}(\mathbf s,t)
+
\gamma_{\tau_h}(\mathbf s,t)
\mathbf v_{\tau_h}^{\star}(\mathbf s,t),
\label{eq:dual_perturbation_velocity_decomposition}
\end{equation}
where
\begin{align}
\gamma_{\tau_\ell}(\mathbf s,t)
&=
\frac{
\omega_\ell p_{\tau_\ell}^{t}(\mathbf s)
}{
\omega_\ell p_{\tau_\ell}^{t}(\mathbf s)
+
\omega_h p_{\tau_h}^{t}(\mathbf s)
},
\label{eq:low_anchor_responsibility}
\\
\gamma_{\tau_h}(\mathbf s,t)
&=
\frac{
\omega_h p_{\tau_h}^{t}(\mathbf s)
}{
\omega_\ell p_{\tau_\ell}^{t}(\mathbf s)
+
\omega_h p_{\tau_h}^{t}(\mathbf s)
}.
\label{eq:high_anchor_responsibility}
\end{align}}
Proof. See Appendix A.
\hfill $\blacksquare$

\noindent\emph{Remark 2:}
Proposition 2 shows that dual-anchor perturbation training yields a shared population-optimal velocity field that adaptively combines the anchor-specific velocities through posterior responsibilities.
These responsibilities act as implicit soft gates over different corruption scales: the low-perturbation anchor dominates near the clean-state manifold, whereas the high-perturbation anchor becomes more influential for severely distorted states.
Therefore, the proposed strategy provides a compact multi-scale restoration mechanism for imperfect-CSI MIMO semantic reception, without requiring a dense perturbation schedule.

\subsection{Cascaded ODE-Based Inference for Restoration Flow Matching}
\label{subsec:ode_based_inference}

Based on the learned restoration vector fields, we develop a cascaded RFM inference scheme by solving two successive recovery ODEs for channel refinement and equalization correction, as illustrated in Fig.~\ref{fig:algorithm_flow}.
An important feature of the proposed inference is that the restoration trajectories are coupled with the physical measurements through observation-dependent initialization. In the channel domain, the ODE starts from the LS estimate, which is directly obtained from the pilot observation and therefore already encodes the pilot measurement constraint. 
Hence, CRFM does not generate a channel realization from an uninformative prior, but performs prior-guided residual correction around a pilot-consistent coarse CSI.
Similarly, in the semantic domain, the initial state is the MMSE-equalized latent representation obtained from the actual data observation using the restored channel, followed by the deterministic receive-side demapping.
This state inherits the regularized data-fitting property of the MIMO receiver chain and serves as a coarse semantic representation. 
This design reduces the risk of the generated distribution being reasonable but inconsistent with measurement results, while retaining the ability to perform rapid ODE inference within a few steps.
Accordingly, fast online inference is achieved with only \(k_{\mathrm H}=5\) CRFM Euler steps and \(k_{\mathrm U}=2\) SRFM Euler steps, as determined by the sensitivity analysis in Section~\ref{subsec:rfm_sensitivity}.

Specifically, the CRFM and SRFM operators are implemented as terminal-state mappings of two deterministic restoration ODEs. 
For the channel-domain trajectory, the initial and terminal states are given by
\(\mathbf H^{[0]}=\mathbf H_{\mathrm{LS}}\) and
\(\widehat{\mathbf H}=\mathbf H^{[k_{\mathrm H}]}\), respectively. 
Using a uniform time grid \(t_{\mathrm H}^{[k]}=k/k_{\mathrm H}\), where \(k_{\mathrm H}\) denotes the number of CRFM sampling steps, and the time step is \(\Delta t_{\mathrm H}=1/k_{\mathrm H}\), the explicit Euler discretization can be expressed as
\begin{equation}
\mathbf H^{[k+1]}
=
\mathbf H^{[k]}
+
\Delta t_{\mathrm H}\,
\mathbf v_{\theta_{\mathrm H}^{\star}}
\left(
\mathbf H^{[k]},
t_{\mathrm H}^{[k]}
\right),
\quad
k=0,\ldots,k_{\mathrm H}-1.
\label{eq:crfm_euler_update}
\end{equation}
After channel restoration, the refined channel \(\widehat{\mathbf H}\) is used
for MMSE equalization according to \eqref{eq:coarse_equalized_signal}, yielding
the coarse semantic latent representation \(\widehat{\mathbf U}_{\mathrm{MMSE}}\).

For the semantic-domain trajectory, the initial and terminal states are
\(\mathbf U^{[0]}=\widehat{\mathbf U}_{\mathrm{MMSE}}\) and
\(\widehat{\mathbf U}=\mathbf U^{[k_{\mathrm U}]}\), respectively.
With \(t_{\mathrm U}^{[k]}=k/k_{\mathrm U}\) and
\(\Delta t_{\mathrm U}=1/k_{\mathrm U}\), the explicit Euler discretization of
the SRFM ODE is given by
\begin{equation}
\mathbf U^{[k+1]}
=
\mathbf U^{[k]}
+
\Delta t_{\mathrm U}\,
\mathbf v_{\theta_{\mathrm U}^{\star}}
\left(
\mathbf U^{[k]},
t_{\mathrm U}^{[k]}
\right),
\quad
k=0,\ldots,k_{\mathrm U}-1.
\label{eq:srfm_euler_update}
\end{equation}

As a result, CRFM first calibrates the pilot-derived LS channel estimate, thereby reducing the CSI mismatch before equalization. SRFM then compensates for the residual latent distortion after MMSE equalization. The restored semantic latent representation \(\widehat{\mathbf U}\) is finally fed into the semantic decoder according to \eqref{eq:semantic_decoder_reconstruction}.

\section{Numerical Results}
\label{Numerical Results}
In this section, we conduct a series of experiments to evaluate the performance of the proposed scheme, providing a comprehensive demonstration of its effectiveness across various scenarios.

\subsection{Experimental Setup}

\subsubsection{Dataset and System Configurations}

The wireless propagation channels are generated according to the CDL-B and CDL-C models specified in 3GPP TR 38.901~\cite{3GPP-CDL}. 
Unless otherwise stated, \(10{,}000\) independent channel realizations are used for training and \(1{,}000\) independent channel realizations are used for testing. 
The carrier frequency is set to \(40\,\mathrm{GHz}\), and the normalized antenna spacing is fixed at \(0.5\). 
Throughout the experiments, we consider a MIMO configuration with \(N_t=64\) transmit antennas and \(N_r=16\) receive antennas. 
The number of spatial streams is set to \(N_s=4\) by default.
For visual semantic transmission, we use two image datasets constructed from ImageNet-mini~\cite{Deng2009ImageNet} and CelebA~\cite{CelebA_2015}. 
All images are center-cropped and resized to \(512\times512\) before semantic encoding. 
The ImageNet-mini split contains \(34{,}745\) training images and \(3{,}923\) testing images, while the CelebA split contains \(2{,}693\) training images and \(300\) testing images.

\subsubsection{Training and Inference Configurations}
All our experiments are performed on a Linux server with RTX 4090 GPU.
The overall framework consists of three learnable components, namely the SwinT-JSCC semantic transceiver, the CRFM module and the SRFM module. 
The SwinT-JSCC encoder-decoder is first trained over an AWGN channel at \(\mathrm{SNR}=10\,\mathrm{dB}\) with a learning rate of \(1\times10^{-4}\). 
After pretraining, the semantic transceiver is frozen, providing fixed latent representations for the SRFM training.
The CRFM and SRFM modules are then trained separately according to the restoration flow matching objectives in \eqref{eq:crfm_dual_anchor_loss} and \eqref{eq:srfm_dual_anchor_loss}, respectively. 
For both modules, the learning rate is set to \(1\times10^{-5}\). 
The dual-anchor perturbation strategy is adopted with \((\tau_{\ell},\tau_{h})=(0.1,1.0)\).
During inference, \(k_{\mathrm H}=5, k_{\mathrm U}=2\) Euler sampling steps are used for CRFM and SRFM, respectively.

\subsubsection{Evaluation Metrics}

For channel estimation, we use the normalized mean square error (NMSE)~\cite{Arvinte_MIMO_2023}, defined as
\begin{equation}
\mathrm{NMSE}\,[\mathrm{dB}]
=
10\log_{10}
\left(
\frac{\|\widehat{\mathbf H}-\mathbf H_{\mathrm e}\|_F^2}
{\|\mathbf H_{\mathrm e}\|_F^2}
\right).
\label{eq:nmse_definition_revised}
\end{equation}
For semantic reconstruction, the channel bandwidth ratio (CBR) is defined as
\begin{equation}
\rho=
\frac{N_c}{C\times H\times W}.
\label{eq:cbr}
\end{equation} 
The CBR was set to \(\rho=1/96\) throughout the experiment.
The reconstruction quality is evaluated using two widely adopted image-quality metrics, namely PSNR and MS-SSIM~\cite{PSNR_SSIM}. 

\subsubsection{Baselines}

We consider two benchmark groups for channel estimation and end-to-end semantic reconstruction. 

For channel estimation, the proposed CRFM is benchmarked against the representative estimators covering classical model-based estimation, sparse recovery, supervised denoising, and generative recovery. 
Specifically, we consider \textbf{LS} estimation~\cite{Ozdemir_Channel_2007} and \textbf{Approximate MMSE}~\cite{Arvinte_MIMO_2023}, \textbf{orthogonal matching pursuit (OMP)}~\cite{Ruan_OMP_NearField_2024} as a sparsity-driven recovery baseline, and \textbf{Score-based method}~\cite{Arvinte_MIMO_2023} and \textbf{DDIM-based method}~\cite{Song_DDIM_2021} as diffusion based generative baselines. 

For semantic reconstruction, we evaluate several ablation configurations of the receiver to quantify the individual and joint effects of CRFM and SRFM.
The AWGN schemes, including \textbf{SwinT-JSCC (train \(10\,\mathrm{dB}\))} and \textbf{SwinT-JSCC with SRFM}, are used as non-MIMO reference cases that only involve AWGN.
For practical imperfect-CSI MIMO transmission, we consider \textbf{LS-MMSE}, \textbf{LS-MMSE with SRFM}, \textbf{LS-MMSE with CRFM}, and \textbf{LS-MMSE with CRFM and SRFM}. \textbf{LS-MMSE} denotes the baseline receiver that uses LS-estimated CSI to construct the MMSE equalizer. 
In addition, \textbf{Perfect-CSI MMSE} and \textbf{Perfect-CSI MMSE with SRFM} are included as upper-bound references to quantify the performance loss caused by pilot-based CSI acquisition. 
A conventional \textbf{JPEG-1/2LDPC-4QAM} system is also evaluated over the same MIMO channel as a separated source-channel coding baseline \cite{Bourtsoulatze_Deep_2019}.

\subsection{Sensitivity Analysis of RFM Design Parameters}
\label{subsec:rfm_sensitivity}
\begin{figure*}[!t]
\centering
\subfloat[]{
\includegraphics[width=0.225\textwidth]{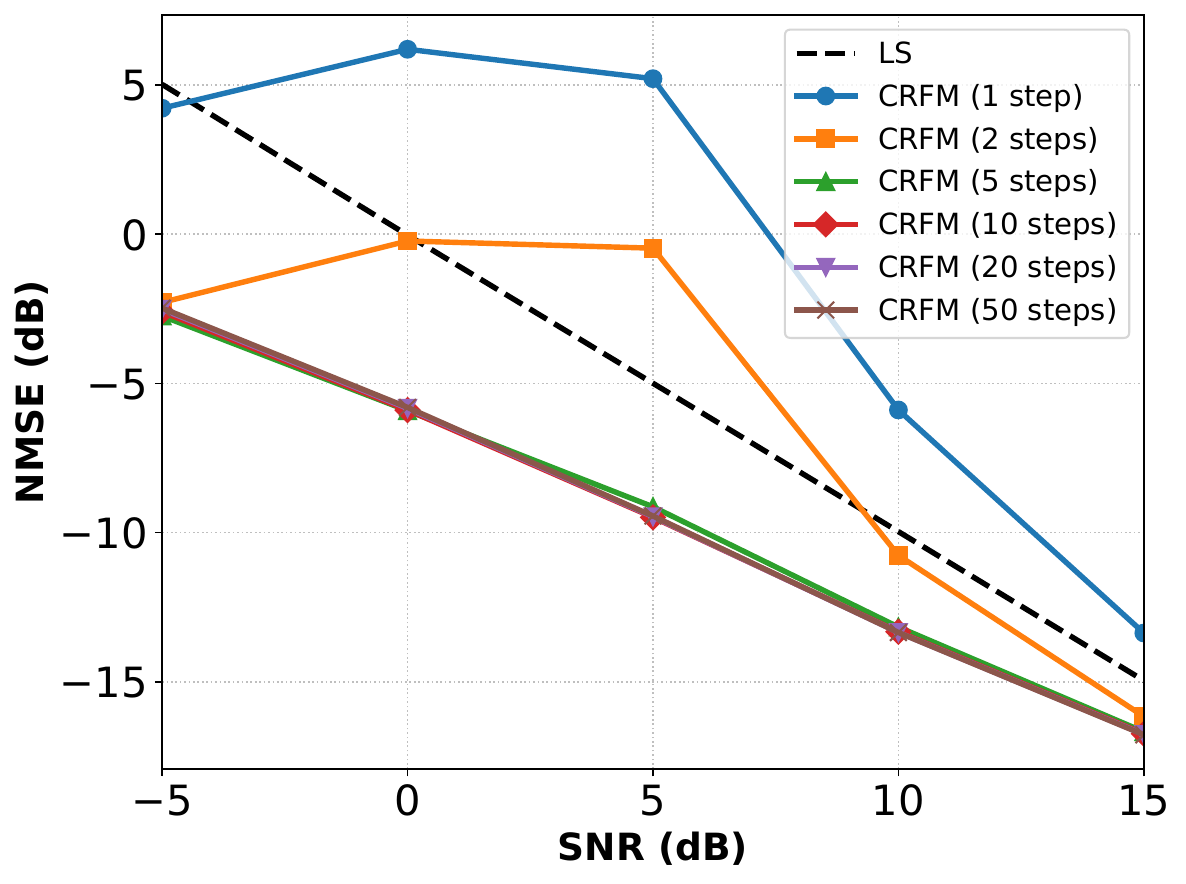}
\label{fig:CDL-C_Channel_diff_steps}
}
\hfil
\subfloat[]{
\includegraphics[width=0.235\textwidth]{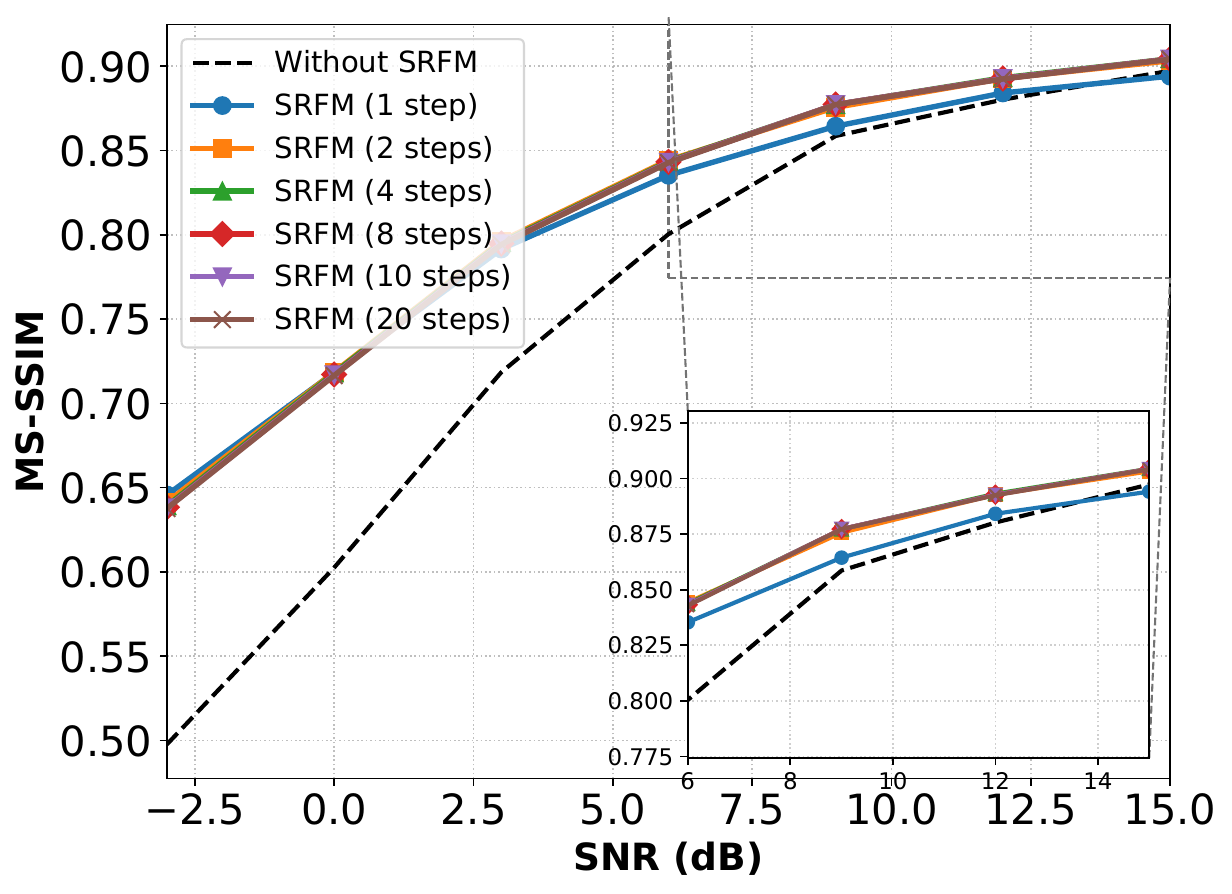}
\label{fig:perfectCSI_msssim_diff_steps}
}
\hfil
\subfloat[]{
\includegraphics[width=0.225\textwidth]{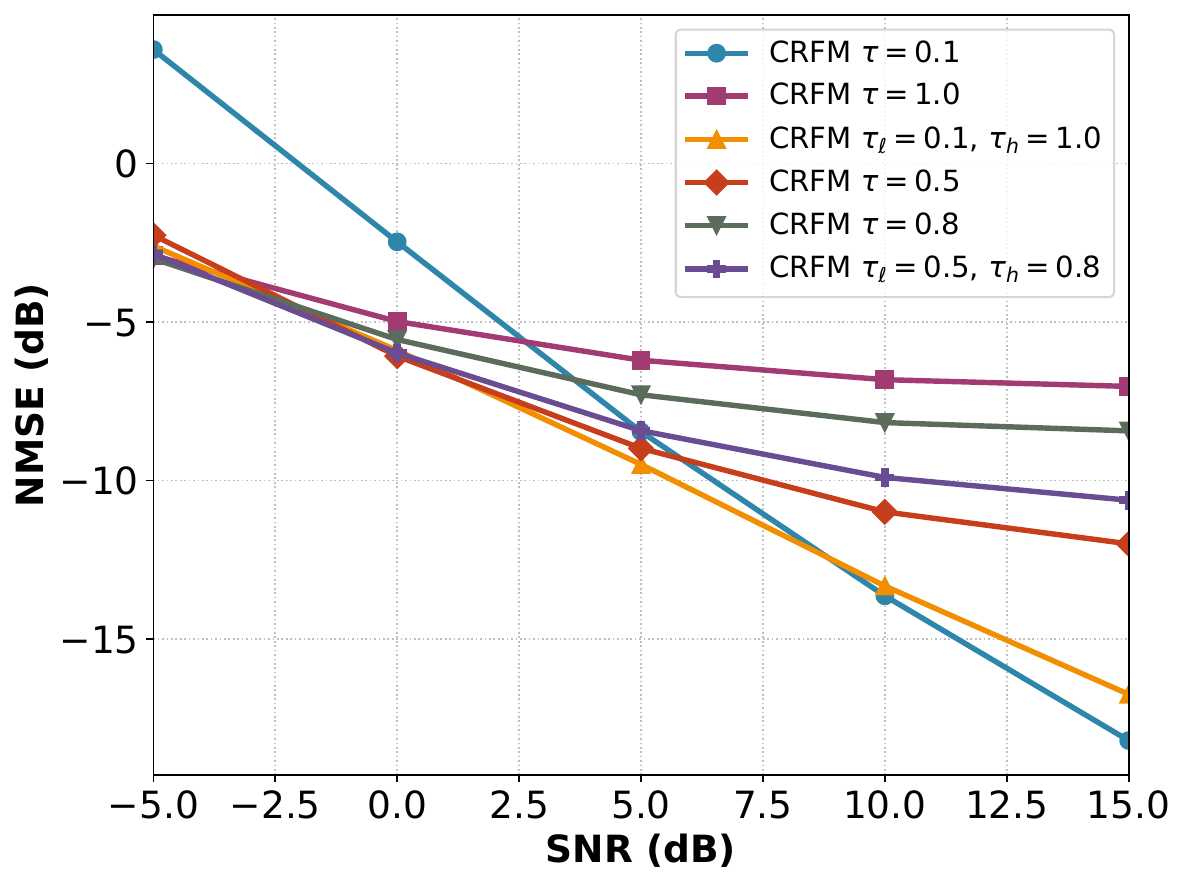}
\label{fig:CDL-C_channel_diff_noise}
}
\hfil
\subfloat[]{
\includegraphics[width=0.225\textwidth]{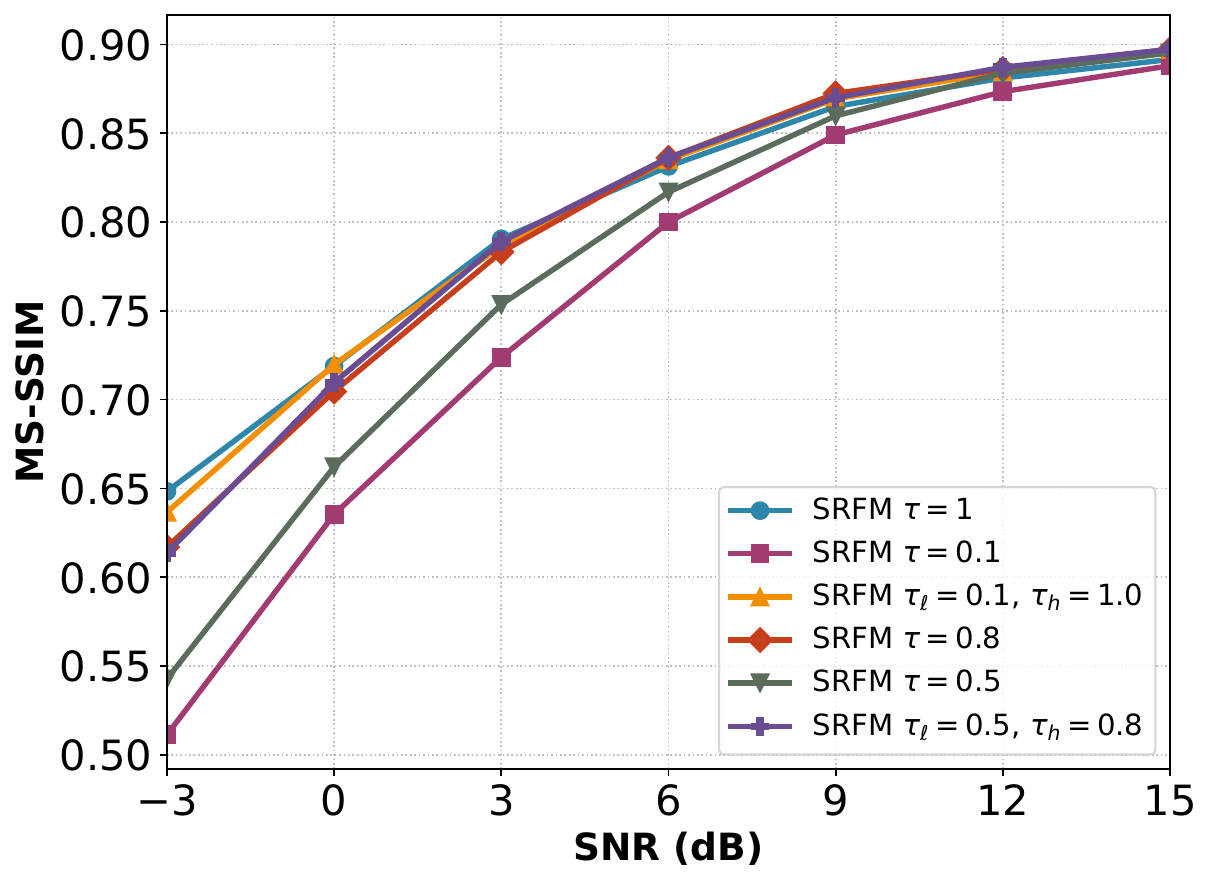}
\label{fig:ImageNet_msssim_noise_level}
}
\caption{Ablation study of the proposed cascaded RFM receiver under different ODE sampling steps and perturbation anchors.
(a) Channel-estimation NMSE versus SNR over the CDL-C channel for different CRFM Euler steps, with $T_p=4$.
(b) Semantic reconstruction MS-SSIM versus SNR for different SRFM Euler steps under perfect CSI.
(c) Channel-estimation NMSE versus SNR over the CDL-C channel for different CRFM perturbation anchors, with $T_p=4$.
(d) Semantic reconstruction MS-SSIM versus SNR for different SRFM perturbation anchors under perfect CSI.}
\label{fig:dfm_steps_noise_results}
\end{figure*}

This subsection examines two key factors of the proposed cascaded RFM receiver: the number of ODE sampling steps and the perturbation anchors used for RFM training.
Fig.~\ref{fig:dfm_steps_noise_results}\subref{fig:CDL-C_Channel_diff_steps} shows the effect of the number of Euler steps on CRFM-based channel restoration. 
A single step is insufficient, especially in the low- and medium-SNR regions, since the learned restoration field cannot be fully integrated. 
As the number of steps increases from $1$ to $2$, the NMSE decreases rapidly, while the gain becomes marginal beyond $10$ steps. 
This indicates that CRFM can reach a near-converged channel estimate with only a few deterministic ODE evaluations. 
Accordingly, we use $k_{\mathrm H}=5$ in the following experiments.
Fig.~\ref{fig:dfm_steps_noise_results}\subref{fig:perfectCSI_msssim_diff_steps} presents the corresponding result for SRFM. 
The MS-SSIM improves noticeably with only a few semantic restoration steps, confirming that SRFM effectively compensates for residual latent distortion after equalization. 
Although more steps can further improve the reconstruction quality, the incremental gain gradually decreases. 
Considering the latency requirement of semantic receivers, we adopt $k_{\mathrm U}=2$ as a practical operating point.
Figs.~\ref{fig:dfm_steps_noise_results}\subref{fig:CDL-C_channel_diff_noise} and \ref{fig:dfm_steps_noise_results}\subref{fig:ImageNet_msssim_noise_level} compare different perturbation anchors. 
The low-perturbation anchor $\tau_{\ell}=0.1$ is more effective in the high-SNR region, where the coarse estimate is already close to the clean manifold. 
By contrast, the high-perturbation anchor $\tau_h=1.0$ provides stronger robustness in the low-SNR region with severe corruption. 
The dual-anchor model with $(\tau_{\ell},\tau_h)=(0.1,1.0)$ achieves the most balanced performance over the whole SNR range.
These results verify that the proposed receiver is both sample-efficient and robust: few-step ODE inference is sufficient for online restoration, while dual-anchor training enables the learned flow to handle both coarse recovery and near-manifold refinement.

\subsection{Evaluation of Channel Estimation Performance}

\begin{figure*}[!t]
\centering
\subfloat[]{
\includegraphics[width=0.235\textwidth]{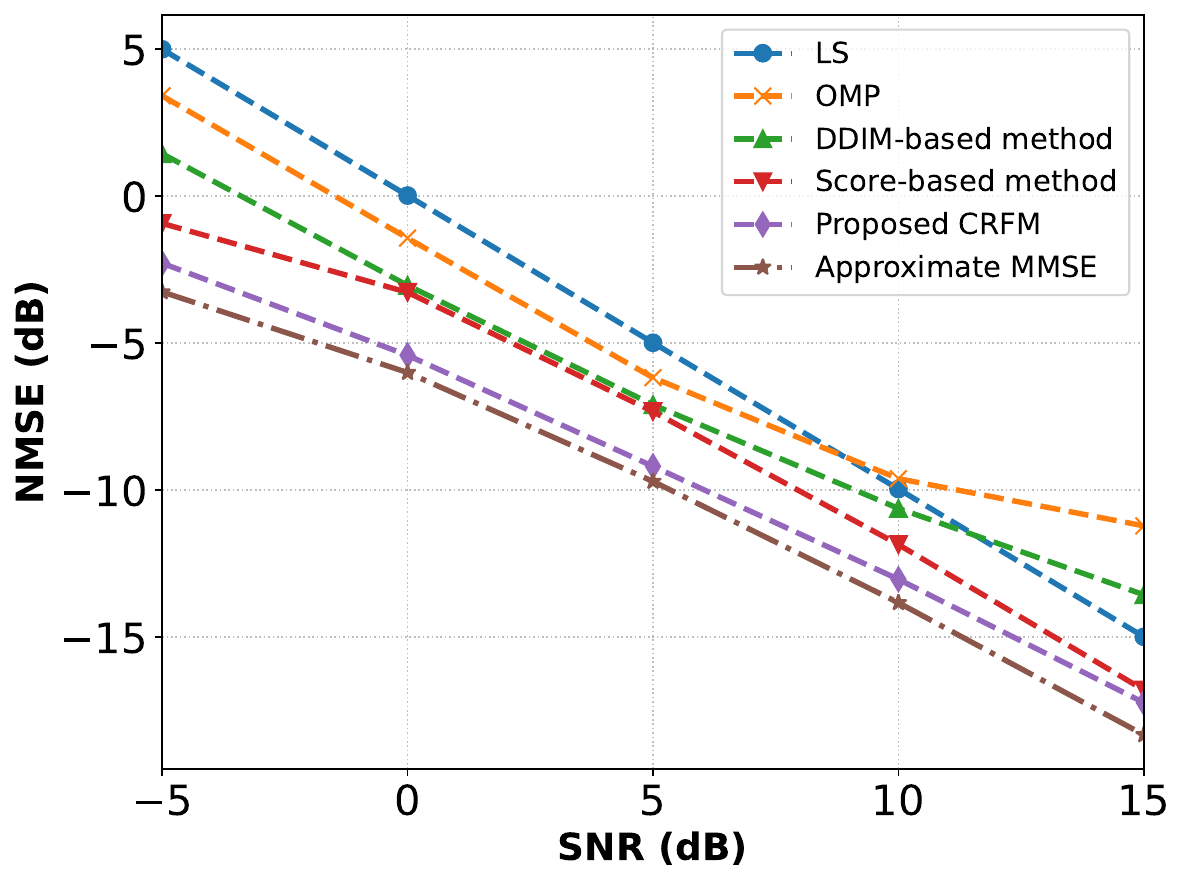}
\label{fig:CDL-B-snr_curve}
}
\hfil
\subfloat[]{
\includegraphics[width=0.235\textwidth]{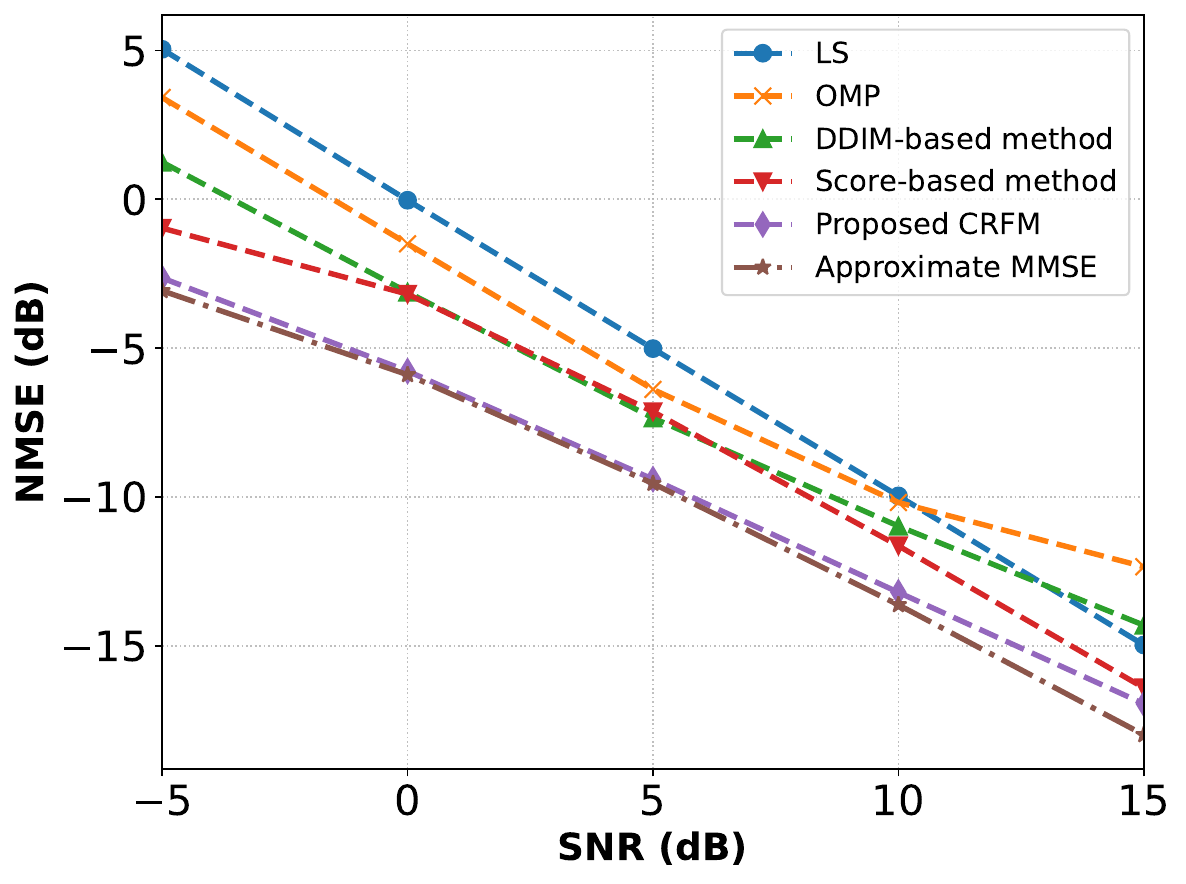}
\label{fig:CDL-C-snr_curve}
}
\hfil
\subfloat[]{
\includegraphics[width=0.230\textwidth]{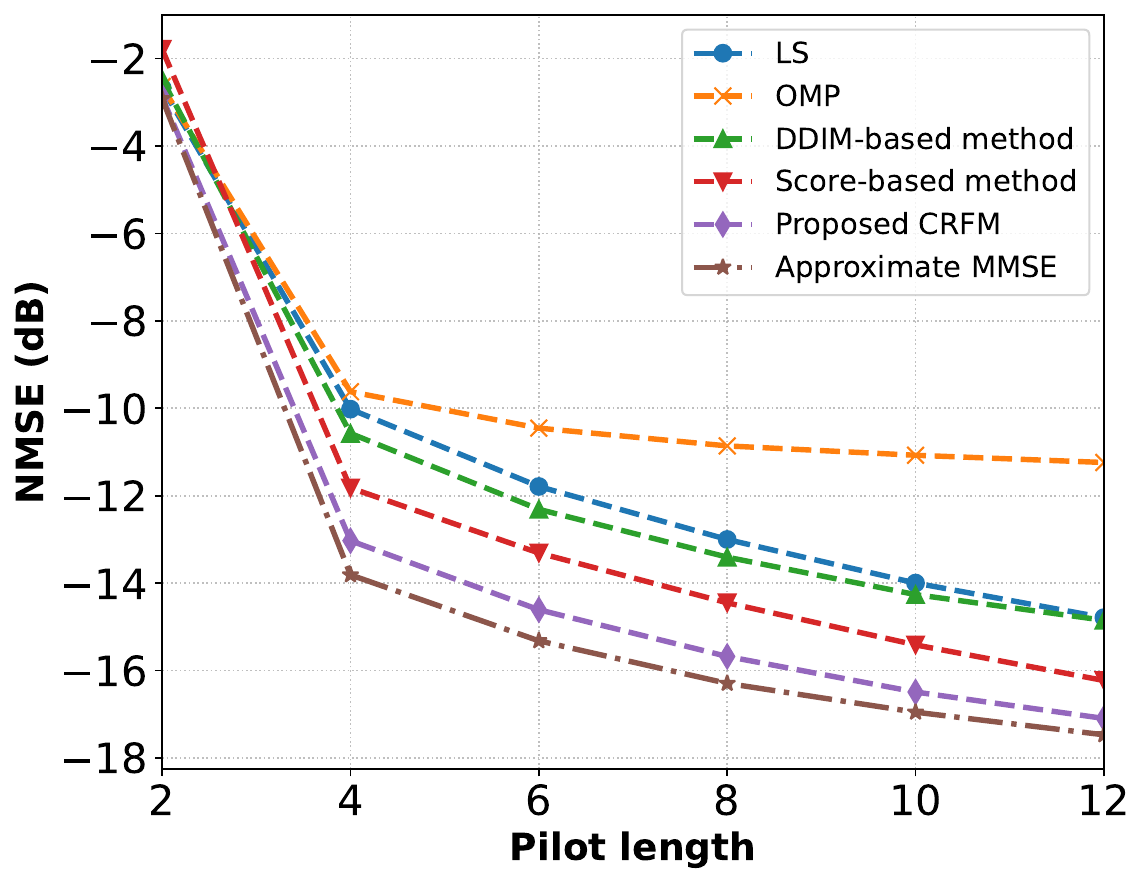}
\label{fig:CDL-B-pilot_curve}
}
\hfil
\subfloat[]{
\includegraphics[width=0.235\textwidth]{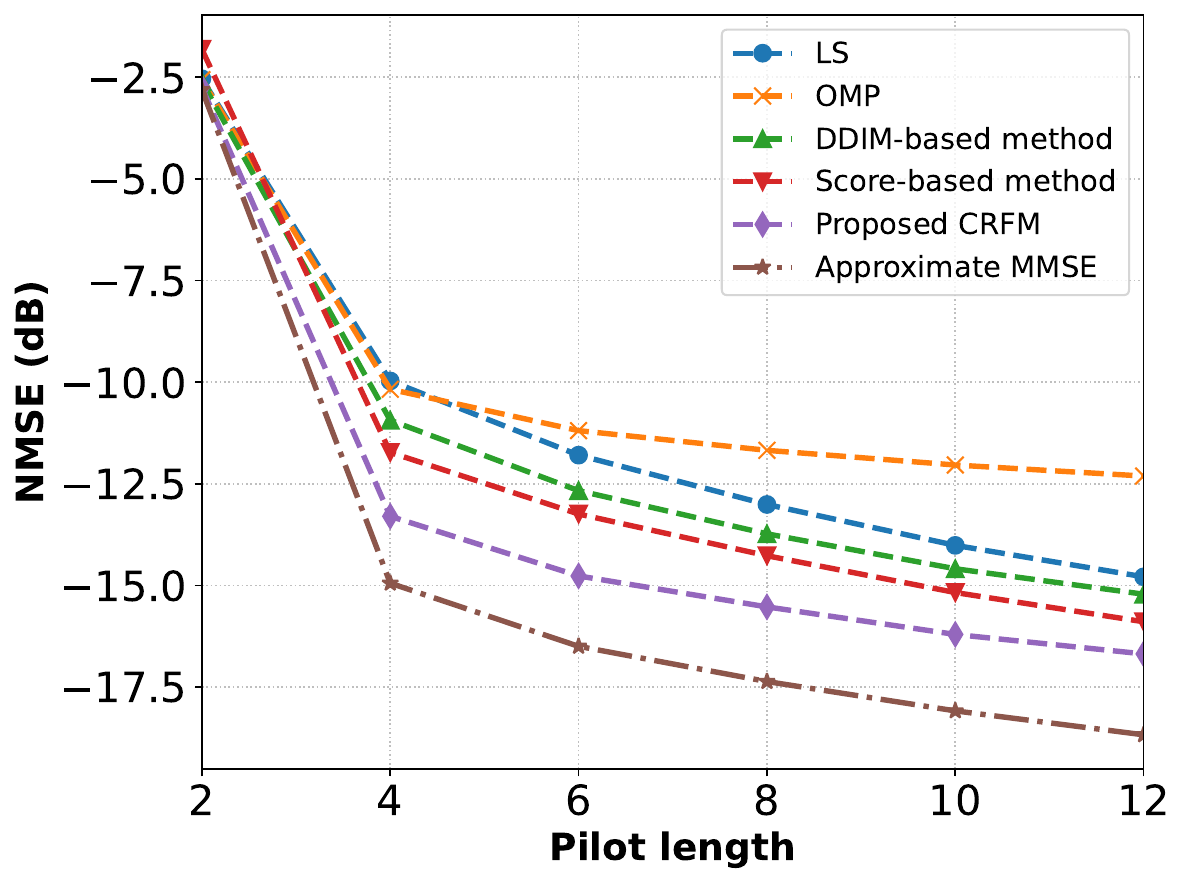}
\label{fig:CDL-C-pilot_curve}
}
\caption{Channel-estimation NMSE comparison over 3GPP CDL-B and CDL-C MIMO channels.
(a) NMSE versus SNR over CDL-B with $T_p=4$.
(b) NMSE versus SNR over CDL-C with $T_p=4$.
(c) NMSE versus pilot length over CDL-B at $\mathrm{SNR}=10\,\mathrm{dB}$.
(d) NMSE versus pilot length over CDL-C at $\mathrm{SNR}=10\,\mathrm{dB}$.}\label{fig:channel_estimation_cdl_results}
\end{figure*}

We next evaluate the channel-estimation performance of the proposed CRFM under different SNRs and pilot lengths. 
The purpose is to verify whether CRFM can improve the coarse LS estimate under both noise-limited and pilot-limited conditions.
Figs.~\ref{fig:channel_estimation_cdl_results}\subref{fig:CDL-B-snr_curve} and \ref{fig:channel_estimation_cdl_results}\subref{fig:CDL-C-snr_curve} show the NMSE versus SNR over the CDL-B and CDL-C channels, respectively, with the pilot length fixed at $T_p=4$. 
The proposed CRFM consistently achieves lower NMSE than LS, OMP, DDIM-based estimation, and score-based estimation over both channel models. 
The gain is especially evident in the low-SNR region, where the LS observation is severely corrupted and conventional sparse or generative baselines suffer from residual estimation errors. 
This confirms that CRFM can effectively learn a coarse-to-clean restoration field from noisy pilot-based channel estimates. 
Compared with the approximate MMSE estimator, CRFM also shows competitive performance without requiring explicit online covariance modeling.
Figs.~\ref{fig:channel_estimation_cdl_results}\subref{fig:CDL-B-pilot_curve} and \ref{fig:channel_estimation_cdl_results}\subref{fig:CDL-C-pilot_curve} further compare the NMSE performance versus pilot length at $\mathrm{SNR}=10\,\mathrm{dB}$. 
As expected, increasing $T_p$ improves the estimation accuracy of all methods, since more pilot observations provide a more reliable coarse channel estimate. 
However, CRFM maintains a clear advantage in the short-pilot regime, where channel recovery is more challenging and pilot overhead is more critical. 
The performance gap gradually decreases as the pilot length increases, but CRFM remains among the best practical schemes over the whole pilot range.

\subsection{The Impact of Different Channel Estimation Algorithms on Semantic Reconstruction}

\begin{figure}[!t]
\centering
\includegraphics[width=2in]{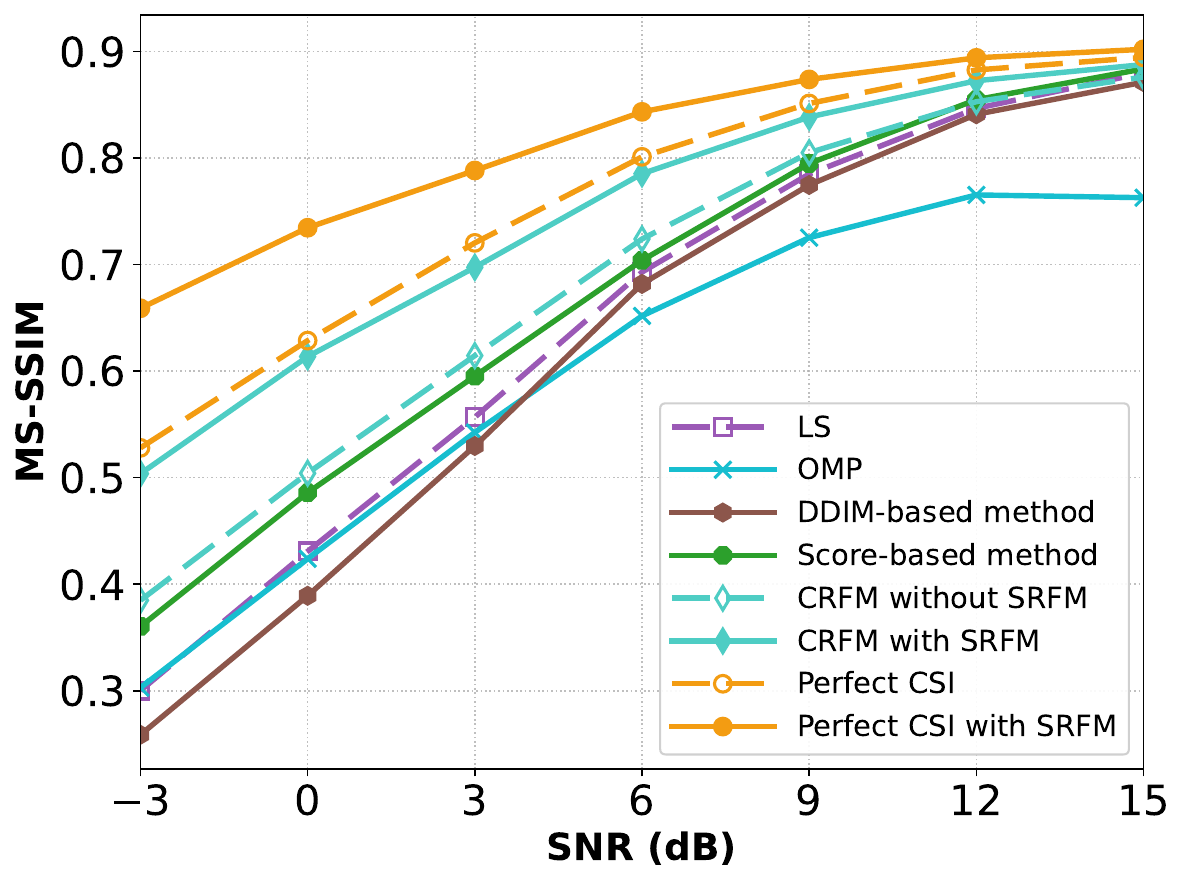}
\caption{Semantic reconstruction MS-SSIM versus SNR under different channel-estimation algorithms, with $T_p=8$ and $\rho=1/96$.}
\label{fig:msssim_diff_CE_algorithm}
\end{figure}

\begin{figure}[!t]
\centering
\includegraphics[width=2in]{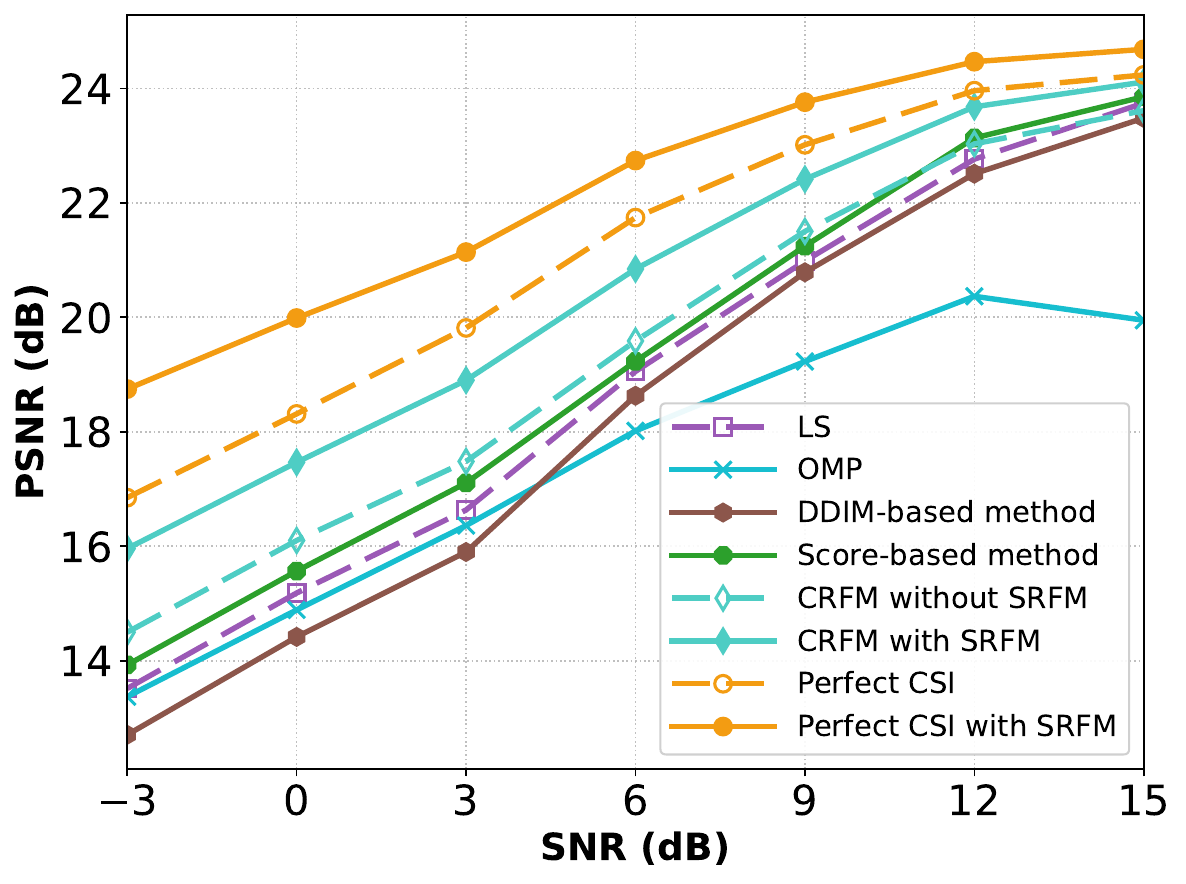}
\caption{Semantic reconstruction PSNR versus SNR under different channel-estimation algorithms, with $T_p=8$ and $\rho=1/96$. }
\label{fig:psnr_diff_CE_algorithm}
\end{figure}

Figs.~\ref{fig:msssim_diff_CE_algorithm} and \ref{fig:psnr_diff_CE_algorithm} show the MS-SSIM and PSNR performance under different channel-estimation algorithms. 
The results reveal a clear correlation between CSI accuracy and semantic reconstruction quality. 
In this experiment, different channel estimators are used to construct the MMSE equalizer, while the semantic encoder, decoder, MIMO configuration, and restoration settings are kept unchanged. 
This comparison isolates the impact of CSI quality on the final image reconstruction performance.
Less accurate channel estimates introduce stronger MMSE equalization mismatch, which appears as residual amplitude-phase distortion and inter-stream interference in the recovered semantic latent representation. 
Consequently, both perceptual similarity and pixel-level fidelity are degraded.
With the proposed CRFM, the receiver achieves consistently better semantic reconstruction over the whole SNR range, indicating that the channel-estimation NMSE gain is effectively transferred to the semantic domain through more reliable equalization. 
Moreover, SRFM further improves the reconstruction quality by refining the residual latent distortion after equalization.

\subsection{Ablation Study of Cascaded CRFM and SRFM for Image Reconstruction}

\begin{figure*}[!t]
\centering
\subfloat[]{
\includegraphics[width=0.235\textwidth]{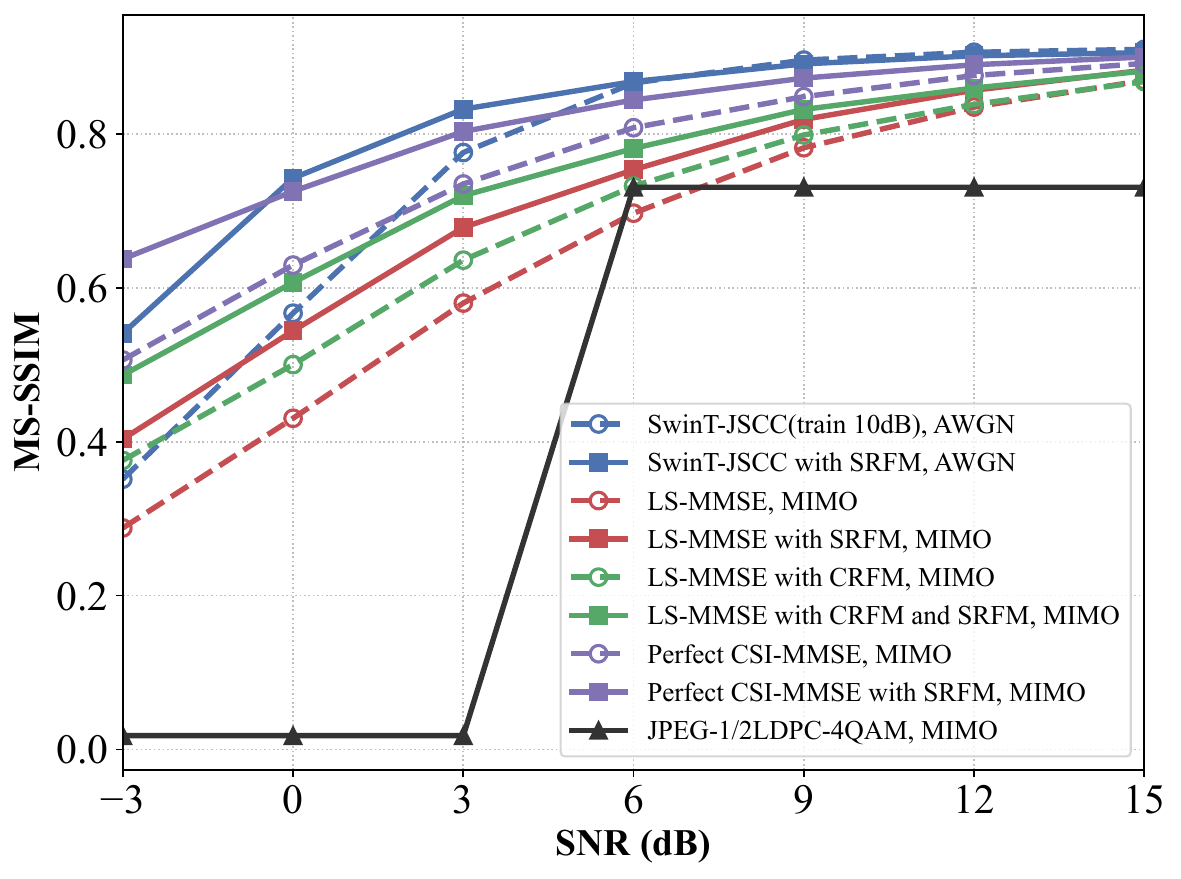}
\label{fig:ImageNet_msssim_CDFM_SDFM}
}
\hfil
\subfloat[]{
\includegraphics[width=0.235\textwidth]{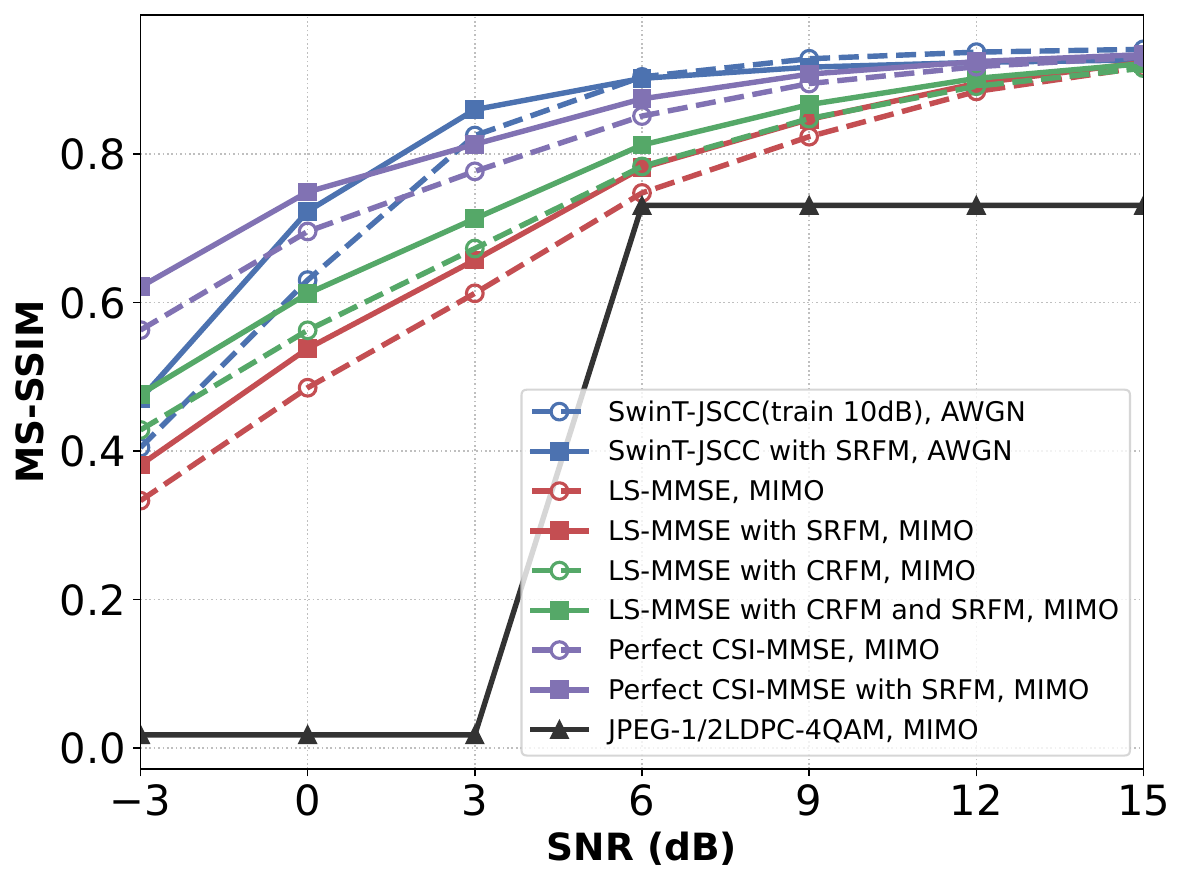}
\label{fig:CelebA_msssim_CDFM_SDFM}
}
\hfil
\subfloat[]{
\includegraphics[width=0.235\textwidth]{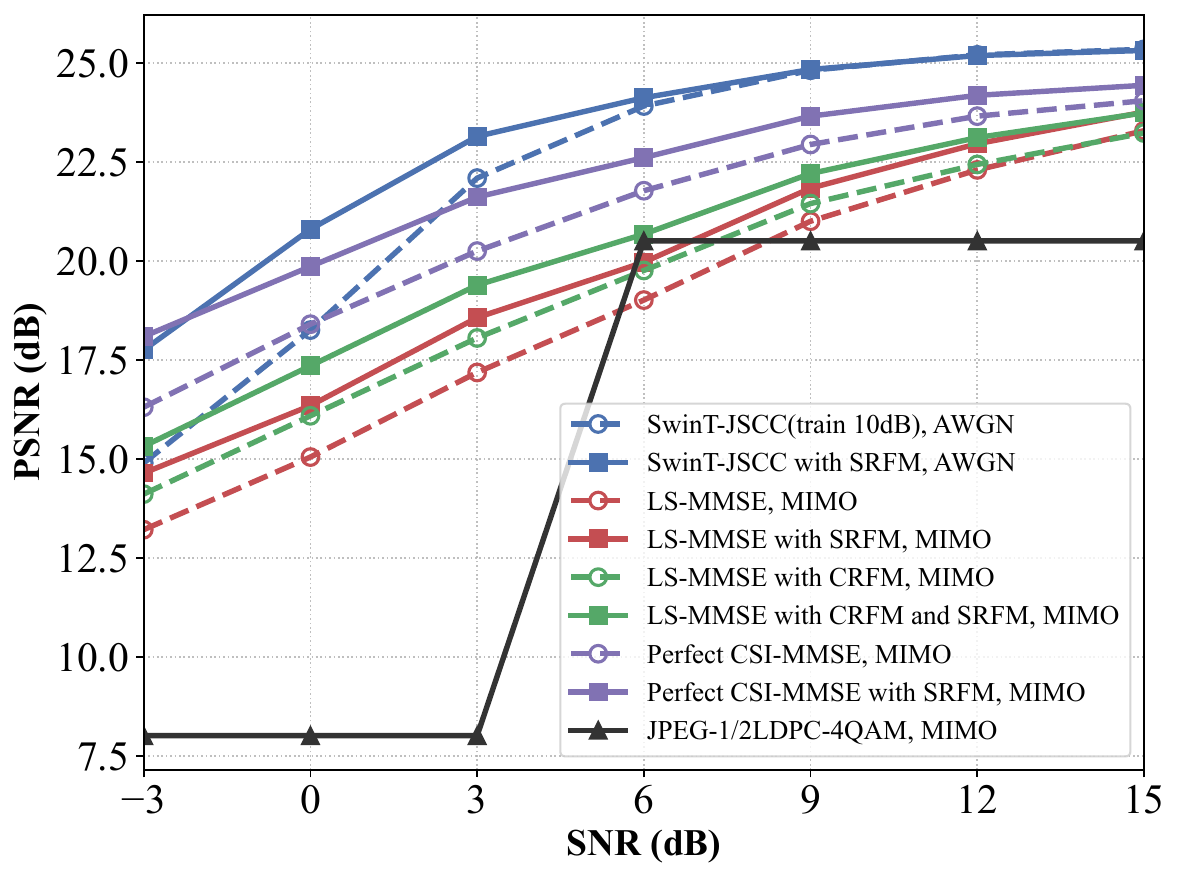}
\label{fig:ImageNet_psnr_CDFM_SDFM}
}
\hfil
\subfloat[]{
\includegraphics[width=0.235\textwidth]{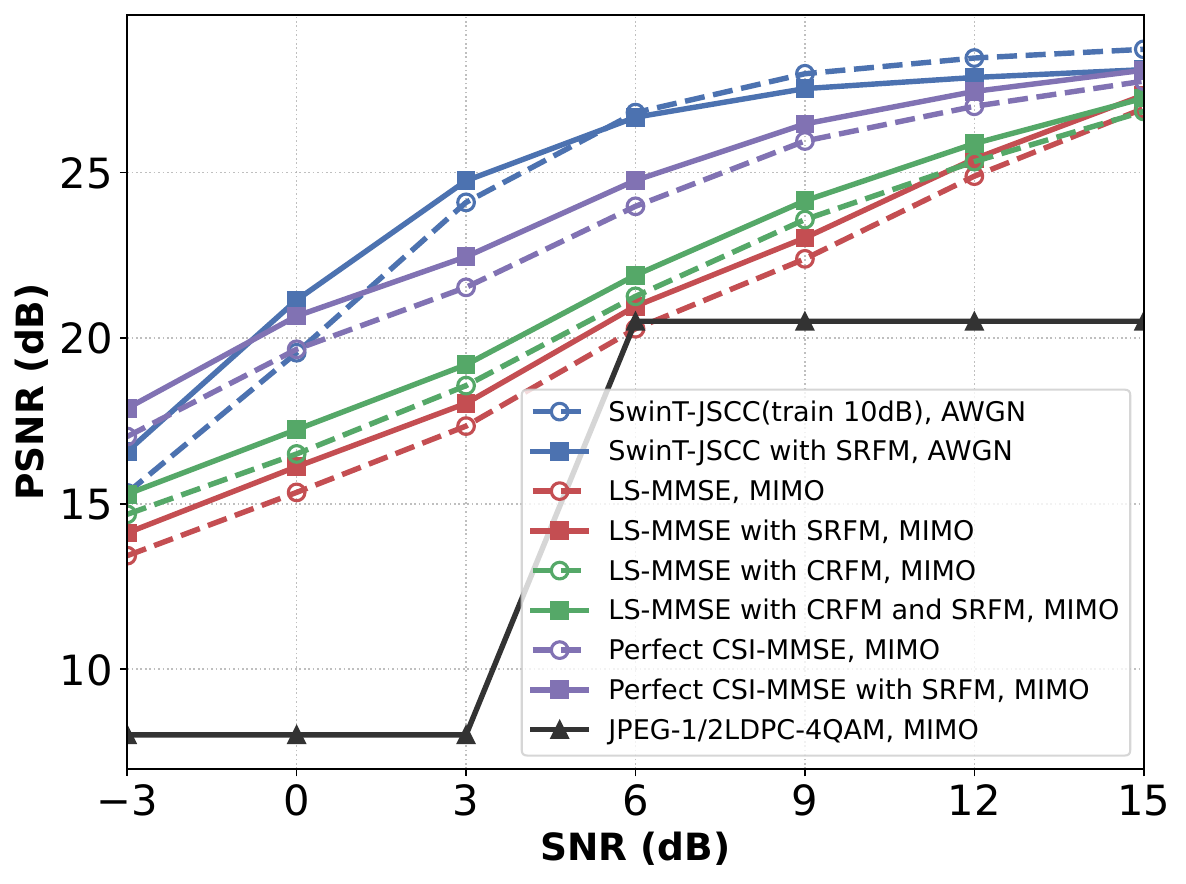}
\label{fig:CelebA_psnr_CDFM_SDFM}
}
\caption{Semantic reconstruction performance under different transmission and receiver schemes on the ImageNet-mini and CelebA datasets, with $T_p=8$ and $\rho=1/96$.
(a) MS-SSIM versus SNR on ImageNet-mini.
(b) MS-SSIM versus SNR on CelebA.
(c) PSNR versus SNR on ImageNet-mini.
(d) PSNR versus SNR on CelebA.} 
\label{fig:semantic_cdfm_sdfm_results}
\end{figure*}

\begin{figure}[!t]
\centering
\includegraphics[width=3.2in]{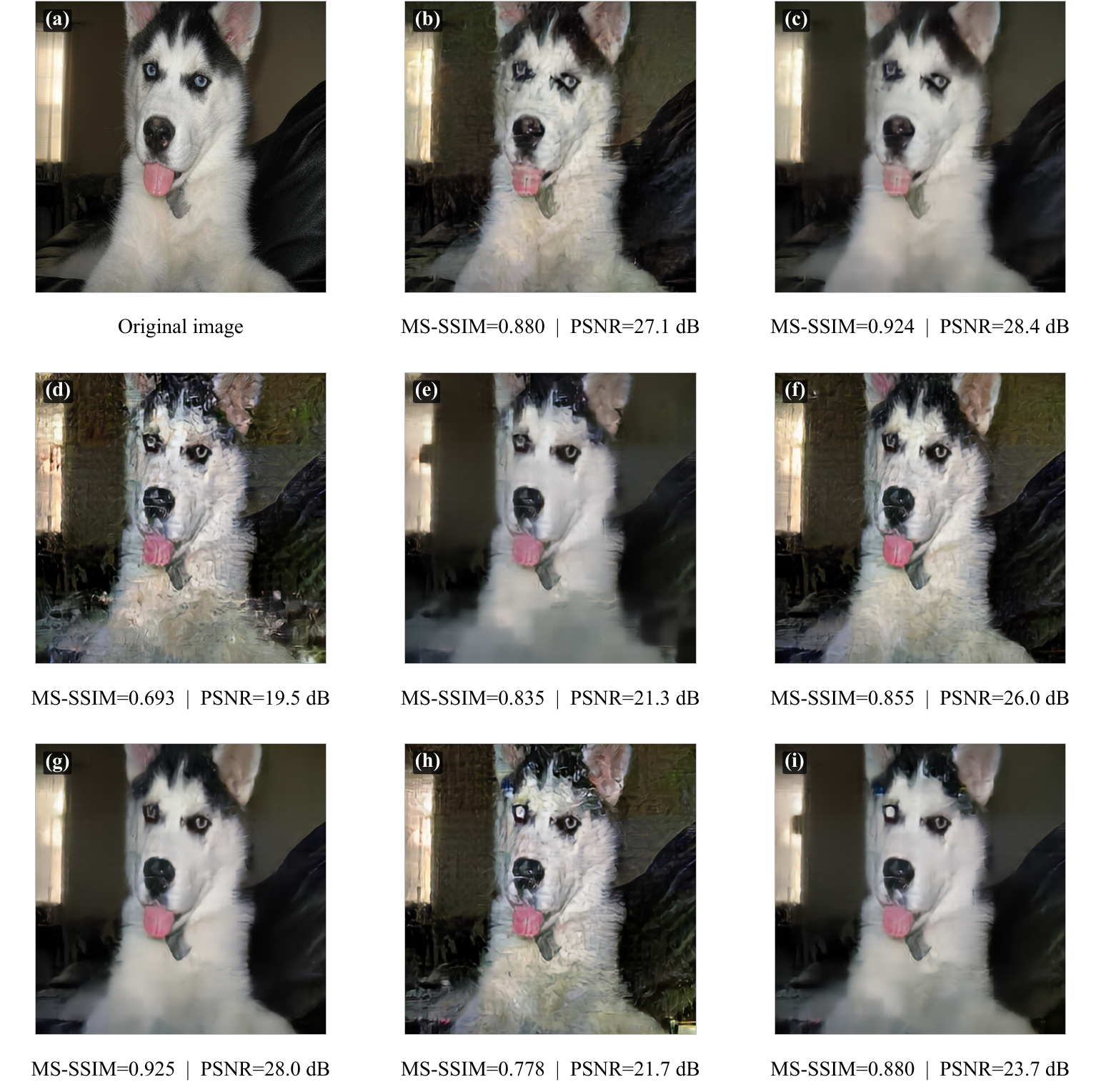}
\caption{Visual reconstruction examples at $\mathrm{SNR}=3\,\mathrm{dB}$. (a) Original image; 
(b) SwinT-JSCC trained at $10\,\mathrm{dB}$ over AWGN; 
(c) SwinT-JSCC with SRFM over AWGN; 
(d) LS-MMSE over MIMO; 
(e) LS-MMSE with SRFM over MIMO; 
(f) perfect-CSI-MMSE over MIMO; 
(g) perfect-CSI-MMSE with SRFM over MIMO. 
(h) LS-MMSE with CRFM over MIMO; 
(i) LS-MMSE with CRFM and SRFM over MIMO; 
The MS-SSIM and PSNR values are reported below each reconstructed image.} 
\label{fig:recon_compare_fixed}
\end{figure}

Fig.~\ref{fig:semantic_cdfm_sdfm_results} evaluates the image reconstruction performance of different receiver configurations on ImageNet-mini and CelebA, including AWGN reference schemes, imperfect-CSI MIMO receivers, perfect-CSI upper-bound receivers, and the JPEG-1/2LDPC-4QAM baseline.
For the AWGN reference case, the degradation mainly comes from additive channel noise and SNR mismatch with the pretrained SwinT-JSCC model. 
The performance gain of AWGN+SRFM over the plain AWGN receiver shows that SRFM can effectively refine semantic latent perturbations caused by additive noise alone. 
In imperfect-CSI MIMO transmission, the LS-MMSE receiver suffers further degradation because pilot-induced channel-estimation errors lead to MMSE equalization mismatch, which introduces structured distortion into the recovered semantic latent representation.
The ablation results show that CRFM and SRFM play complementary roles. 
Therefore, the cascaded CRFM-SRFM receiver achieves the best performance among all imperfect-CSI schemes and substantially approaches the perfect-CSI upper bound over a wide SNR range. 
The consistent improvements on both datasets demonstrate the effectiveness of the proposed cascaded restoration design in enhancing perceptual similarity and pixel-level fidelity. 
In contrast, the JPEG-1/2LDPC-4QAM baseline exhibits a typical low-SNR cliff effect, whereas the JSCC-based schemes degrade more gracefully.
Fig.~\ref{fig:recon_compare_fixed} provides representative visual examples. 
The LS-MMSE receiver produces visible artifacts and structural distortions, whereas CRFM alleviates channel-induced degradation and SRFM further improves semantic details and textures. 
The cascaded CRFM-SRFM receiver yields the most faithful reconstruction among the imperfect-CSI schemes, which is consistent with the quantitative results in Fig.~\ref{fig:semantic_cdfm_sdfm_results}.

\begin{table}[!t]
\centering
\caption{Computational complexity and inference latency comparison.} \label{tab:macs_latency}
\scriptsize
\setlength{\tabcolsep}{8pt}
\renewcommand{\arraystretch}{1}
\begin{tabular}{lccc}
\toprule
\textbf{Method} & \textbf{MACs/Forward} & \textbf{Params} & \textbf{Latency}\\
\midrule

Score-based method
& $40.98$M 
& $2.68$M 
& $6.58$s\\

DDIM-based method
& $41.05$M 
& $2.68$M 
& $831.19$ms \\

CRFM 
& $40.98$M 
& $2.68$M 
& $0.52$ms \\

SRFM 
& $21.8$G
& $129.81$M 
& $35.78$ms \\
\bottomrule

\end{tabular}
\end{table}
\subsection{Complexity and Latency Comparison}

To evaluate the online receiver overhead from an implementation perspective, Table~\ref{tab:macs_latency} reports the multiply--accumulate operations (MACs), the number of trainable parameters, and the inference latency of the considered receiver-side modules.
MACs/Forward denotes the computational cost of one neural-network evaluation, whereas Latency denotes the average runtime of the complete inference procedure under the adopted sampling configuration.
For a fair comparison,the score-based method, the DDIM-based method, and CRFM are implemented with the same UNet-based network backbone \cite{Ho_DDPM_2020}.
Therefore, these three methods have nearly identical single-forward computational costs and model sizes, with about  \(41\)M MACs and \(2.68\)M trainable parameters. 
Their latency difference mainly comes from the number and type of sampling steps. Specifically, the latency of the score-based method includes the complete sampling chain with \(50\) noise levels and \(3\) Langevin correction steps at each level, resulting in \(50\times 3\) score-network evaluations. The latency of the DDIM-based method includes the full \(20\)-step reverse sampling chain. 
In contrast, CRFM performs deterministic ODE inference with only \(k_{\mathrm H}=5\) Euler steps.
Under the measured setting, CRFM achieves the lowest infercence latency among the compared generative channel estimators.
The semantic restoration module SRFM has a larger model size than CRFM, with \(21.8\)G MACs and \(129.81\)M trainable parameters.
This is expected because SRFM operates on the high-dimensional semantic latent representation rather than the low-dimensional channel tensor.

\section{Conclusion}
\label{Conclusion}

This paper investigated MIMO semantic communications under imperfect-CSI and formulated the receiver-side processing as two cascaded inverse recovery problems. 
To address the error propagation from channel estimation to semantic reconstruction, we proposed a cascaded restoration flow matching receiver consisting of CRFM and SRFM. 
CRFM refines the coarse LS channel estimate before MMSE equalization, while SRFM restores the residual distortion in the equalized semantic latent representation. 
Simulation results over MIMO channels and visual semantic transmission tasks showed that the proposed method improves both physical-layer channel estimation and image reconstruction. 
The results confirm that channel restoration and semantic latent restoration are complementary for robust MIMO semantic reception. 
Future work will extend the proposed framework to time-varying and multi-user MIMO scenarios, and further explore adaptive sampling strategies and multimodal semantic transmission.

\section*{Appendix A}
By the $L^2$ projection property of conditional expectation, the population minimizer of
\eqref{eq:dual_perturbation_population_regression} is given by
\begin{equation}
\mathbf v_{\tau_\ell,\tau_h}^{\star}(\mathbf s,t)
=
\mathbb E_{\mu_{\tau_\ell,\tau_h}^{t}}
\left[
\mathbf u
\,\middle|\,
\mathbf S=\mathbf s
\right].
\label{eq:proof_conditional_expectation}
\end{equation}
Using the mixture law in \eqref{eq:dual_perturbation_joint_law}, the above conditional expectation can be decomposed as
\begin{align}
&\mathbb E_{\mu_{\tau_\ell,\tau_h}^{t}}
\left[
\mathbf u
\,\middle|\,
\mathbf S=\mathbf s
\right]
\nonumber\\
&=
\frac{
\omega_\ell p_{\tau_\ell}^{t}(\mathbf s)
}{
p_{\tau_\ell,\tau_h}^{t}(\mathbf s)
}
\mathbb E_{\mu_{\tau_\ell}^{t}}
\left[
\mathbf u
\,\middle|\,
\mathbf S=\mathbf s
\right]
\nonumber\\
&\quad+
\frac{
\omega_h p_{\tau_h}^{t}(\mathbf s)
}{
p_{\tau_\ell,\tau_h}^{t}(\mathbf s)
}
\mathbb E_{\mu_{\tau_h}^{t}}
\left[
\mathbf u
\,\middle|\,
\mathbf S=\mathbf s
\right].
\label{eq:proof_mixture_conditional_expectation}
\end{align}
From the marginal mixture density in
\eqref{eq:dual_perturbation_marginal_density}, we have
\begin{equation}
p_{\tau_\ell,\tau_h}^{t}(\mathbf s)
=
\omega_\ell p_{\tau_\ell}^{t}(\mathbf s)
+
\omega_h p_{\tau_h}^{t}(\mathbf s).
\label{eq:proof_mixture_marginal_density}
\end{equation}
Moreover, by the anchor-wise optimal velocity definition in
\eqref{eq:anchor_velocity_field_compact},
\begin{equation}
\mathbb E_{\mu_{\tau_\ell}^{t}}
\left[
\mathbf u
\,\middle|\,
\mathbf S=\mathbf s
\right]
=
\mathbf v_{\tau_\ell}^{\star}(\mathbf s,t),
\qquad
\mathbb E_{\mu_{\tau_h}^{t}}
\left[
\mathbf u
\,\middle|\,
\mathbf S=\mathbf s
\right]
=
\mathbf v_{\tau_h}^{\star}(\mathbf s,t).
\label{eq:proof_anchor_velocity_definitions}
\end{equation}
Substituting \eqref{eq:proof_mixture_marginal_density} and
\eqref{eq:proof_anchor_velocity_definitions} into
\eqref{eq:proof_mixture_conditional_expectation} yields
\begin{align}
\mathbf v_{\tau_\ell,\tau_h}^{\star}(\mathbf s,t)
&=
\frac{
\omega_\ell p_{\tau_\ell}^{t}(\mathbf s)
}{
\omega_\ell p_{\tau_\ell}^{t}(\mathbf s)
+
\omega_h p_{\tau_h}^{t}(\mathbf s)
}
\mathbf v_{\tau_\ell}^{\star}(\mathbf s,t)
\nonumber\\
&\quad+
\frac{
\omega_h p_{\tau_h}^{t}(\mathbf s)
}{
\omega_\ell p_{\tau_\ell}^{t}(\mathbf s)
+
\omega_h p_{\tau_h}^{t}(\mathbf s)
}
\mathbf v_{\tau_h}^{\star}(\mathbf s,t).
\label{eq:proof_velocity_decomposition}
\end{align}
Recognizing the two coefficients in
\eqref{eq:proof_velocity_decomposition} as
$\gamma_{\tau_\ell}(\mathbf s,t)$ and
$\gamma_{\tau_h}(\mathbf s,t)$ defined in
\eqref{eq:low_anchor_responsibility} and
\eqref{eq:high_anchor_responsibility}, respectively, we obtain
\eqref{eq:dual_perturbation_velocity_decomposition}.
This completes the proof.
\hfill $\blacksquare$

\end{document}